%% file: arxiv.tex
\documentclass[letterpaper, 10pt, conference]{ieeeconf}  % Comment this line out if you need a4paper
\makeatletter
\let\NAT@parse\undefined
\makeatother
\usepackage{times}
\usepackage{array}
\usepackage{graphicx}
\usepackage{amsmath}
\usepackage{amssymb}
\usepackage{tikz}
\usepackage{comment}
\usepackage{wrapfig}
\usepackage{algorithmic}
\usepackage[ruled]{algorithm2e}
\usepackage[labelfont={color=black}]{caption}

\IEEEoverridecommandlockouts                              % This command is only needed if 
\usepackage[square,numbers]{natbib}
\usepackage{multicol}
\usepackage[bookmarks=true]{hyperref}
\definecolor{darkblue}{rgb}{0.0, 0.0, 0.55}
\hypersetup{
    colorlinks=true,
    linkcolor=blue,
    filecolor=blue,
    urlcolor=darkblue,
    citecolor=blue,
    urlbordercolor=white,
}

\let\oldurl\url
\renewcommand{\url}[1]{\textcolor{darkblue}{\small\fontseries{sb}\selectfont\texttt{\oldurl{#1}}}}

\begin{document}

% paper title
%\title{On the Effectiveness of Retrieval, Alignment and Replay in Manipulation}

\title{\LARGE \bf
Adapting Skills to Novel Grasps: A Self-Supervised Approach
}
% \title{Adapting Skills to Novel Grasps: \\A Self-Supervised Approach}

\author{Georgios Papagiannis$^{*}$, Kamil Dreczkowski, Vitalis Vosylius and Edward Johns\\\vspace{-1.5ex}\\The\textbf{ Robot Learning Lab} at \textbf{Imperial College London}\\\vspace{-2.2ex}\\\href{https://www.robot-learning.uk/adapting-skills}{\textbf{www.robot-learning.uk/adapting-skills}}
\thanks{$^{*}$Contact at:
        {\tt\small g.papagiannis21@imperial.ac.uk}}
        }

\maketitle

\begin{abstract}
In this paper, we study the problem of adapting manipulation trajectories involving grasped objects (e.g. tools) defined for a \textit{single} grasp pose to \textit{novel} grasp poses. A common approach to address this is to define a new trajectory for each possible grasp explicitly, but this is highly inefficient. Instead, we propose a method to adapt such trajectories directly while only requiring a period of self-supervised data collection, during which a camera observes the robot's end-effector moving with the object rigidly grasped. Importantly, our method requires no prior knowledge of the grasped object (such as a 3D CAD model), it can work with RGB images, depth images, or both, and it requires no camera calibration. Through a series of real-world experiments involving 1360 evaluations, we find that self-supervised RGB data consistently outperforms alternatives that rely on depth images including several state-of-the-art pose estimation methods. Compared to the best-performing baseline, our method results in an average of $\text{\textbf{28.5}}\%$ higher success rate when adapting manipulation trajectories to novel grasps on several everyday tasks. Videos of the experiments are available on our webpage at \href{https://www.robot-learning.uk/adapting-skills}{www.robot-learning.uk/adapting-skills}.

\end{abstract}
\IEEEpeerreviewmaketitle

% \begin{refsection}

\section{Introduction}

\begin{figure}[t!]
    \centering
\includegraphics[width=.49\textwidth]{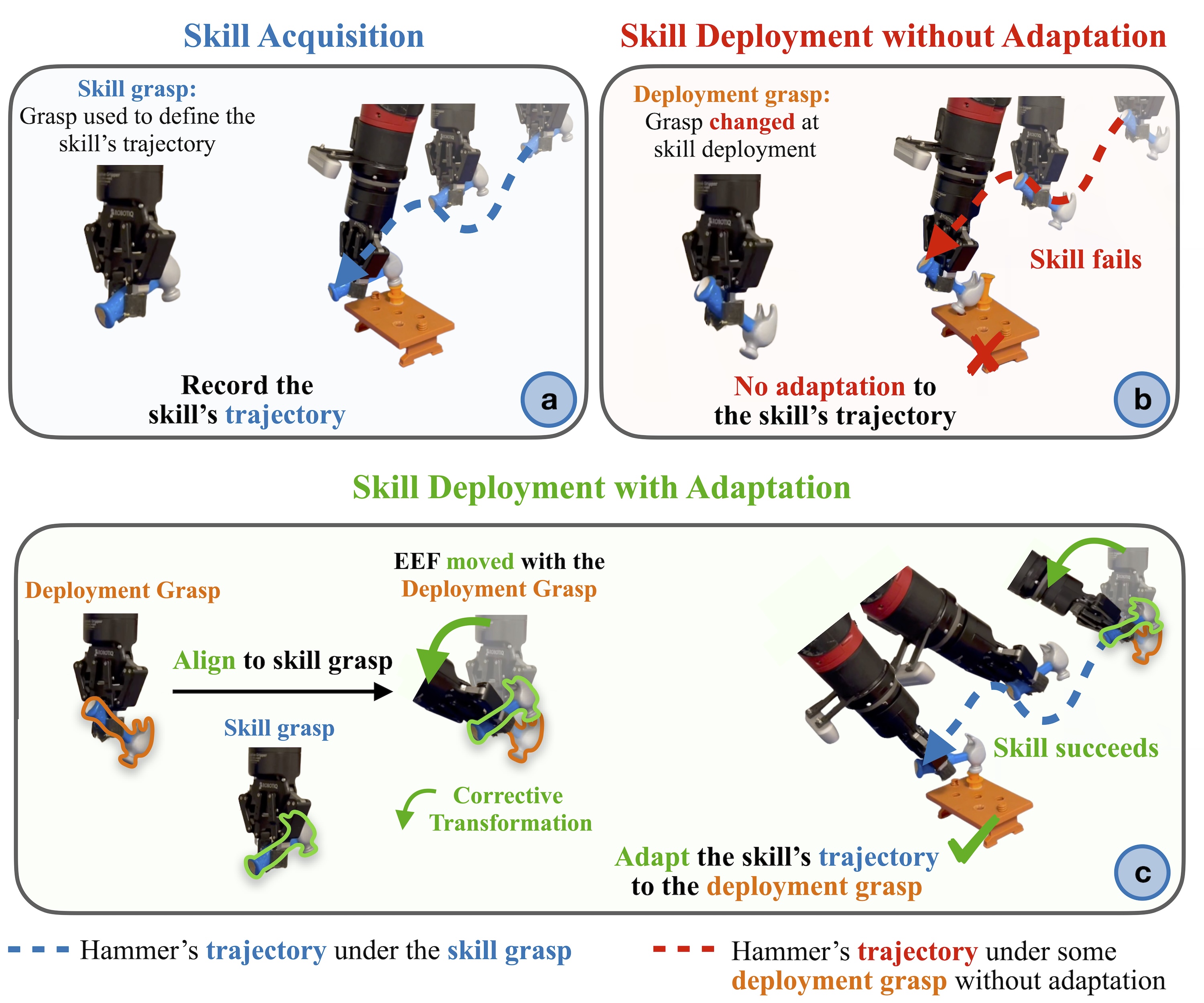}
    \vspace{-.37cm}\caption{\textit{Skill Acquisition:} (a) With the hammer grasped at the skill grasp, the skill's trajectory (defined e.g. via a human demonstration) specifies how to hammer the nail. \textit{Skill Deployment without Adaptation:} (b) If the hammer is grasped differently to the skill grasp, executing the skill's trajectory leads to task failure. \textit{Skill Deployment with Adaptation:} (c) A corrective transformation is applied to the skill's EEF trajectory such that (without changing the grasp pose) the hammer under the deployment grasp follows the same trajectory it followed under the skill grasp, to successfully complete the task.}\vspace{-4.8ex}
    \label{fig:front-page}
\end{figure}

Consider a robot that has acquired a skill involving a grasped object, such as using a hammer to hammer in a nail, as shown in Figure \ref{fig:front-page} (a). Such a skill can be defined with various methods, such as imitation learning \cite{valassakis2022dome}, and comprises a trajectory of end-effector (EEF) poses, and as a result, a trajectory of poses followed by the grasped object. For that skill to be widely applicable, it must generalise to the novel grasp poses that may occur at skill deployment (i.e. to different poses of the hammer within the robot's gripper). However, generalising a skill to novel grasp poses typically requires manually defining a new trajectory for each grasp pose (e.g. by providing multiple grasp-specific demonstrations), which is highly laborious. In this work, we address this problem and develop a method that enables a robot to autonomously adapt a skill's trajectory defined for an object grasped at a single grasp pose, to any novel grasp.

Past work has addressed this problem mainly by regrasping the object at a pose that aligns with the skill's requirements \cite{weiwei2019regrasp, Nguyen2016preparatory, cheng2021learning2regrasp}. However, the majority of these methods assume prior knowledge about the object's 3D CAD model \cite{weiwei2019regrasp, Nguyen2016preparatory} which is usually unavailable in unstructured environments, and methods that bypass this requirement rely on depth images that can cause failures due to missing depth data \cite{cheng2021learning2regrasp}. Moreover, regrasping methods often rely on a sequence of pick and place operations making them slow to deploy \cite{cheng2021learning2regrasp}.

% And while \cite{zengyi2020keto} directly grasps an object in a way that aligns with a task's requirements, it requires prior task-specific training data which may not be readily available.

% Past approaches mainly involve object regrasping \cite{weiwei2019regrasp, Nguyen2016preparatory, cheng2021learning2regrasp} or task-oriented grasping \cite{zengyi2020keto}, aiming to align the grasp with task requirements. Most methods assume prior knowledge of the object's 3D CAD model \cite{weiwei2019regrasp, Nguyen2016preparatory}, often unavailable in unstructured environments. Although \cite{cheng2021learning2regrasp} bypasses this requirement, it relies on depth images, which can fail due to missing data. Moreover, regrasping methods can be slow and depend on stable object placements, leading to failures \cite{cheng2021learning2regrasp}. Task-oriented grasping \cite{zengyi2020keto} requires prior task-specific training data, which may not always be available.

Instead, we are interested in methods that can adapt skills {immediately}, without regrasping or any prior object knowledge. To adapt a skill's trajectory to some grasp without regrasping, a potential solution is to determine a \textit{corrective transformation} that changes the EEF's trajectory during skill deployment, such that the grasped object follows the same trajectory it followed under the grasp pose that was used to define the skill, as shown in Figure~\ref{fig:front-page} (a) and (c). This can be achieved with the use of existing pose estimation methods, to first estimate the relative pose of the grasped object between the skill acquisition and deployment phases, using RGB or depth images of that object captured from a camera. Then, by using the estimated relative pose, the corrective transformation can be obtained trivially using knowledge of the camera's extrinsics \cite{simeonovdu2021ndf}. Given images of the skill and deployment grasps, the required pose estimation could be achieved by establishing correspondences directly \cite{chen2022aspanformer, ICP}, though learned descriptors \cite{ amir2021deep, florence2018dense, simeonovdu2021ndf} or canonical object representations \cite{Wen2022YouOD}, or by explicitly estimating an object's pose \cite{pmlr-v205-goodwin23a, deng2020self-supervised, li2019category, Devgon2020OrientingN3, he2022fs6d}. However, similarly to regrasping approaches, the majority of these methods rely on prior knowledge of the object's 3D CAD model \cite{Wen2022YouOD, deng2020self-supervised} or semantic category-specific training data \cite{simeonovdu2021ndf, Wen2022YouOD, li2019category}, which may not be readily available. And methods that assume no prior object knowledge when learning descriptors \cite{florence2018dense, amir2021deep, pmlr-v205-goodwin23a} or estimating an object's pose \cite{pmlr-v205-goodwin23a, Devgon2020OrientingN3, he2022fs6d, ICP} still rely on depth images, which can be problematic due to noisy or partially missing depth data \cite{Kadambi2014}. Finally, as these methods rely on estimating the pose of the grasped object, to determine the corrective transformation they require precise knowledge of a camera's extrinsic parameters (relative to the EEF), which can negatively affect performance due to challenges with camera calibration \cite{valassakis2021learning}.

\textbf{Our contributions.} Motivated by the above challenges, this paper develops a method to address them by \textbf{bypassing the need to explicitly estimate the pose of a grasped object}. Instead, it \textbf{directly determines the corrective transformation} to adapt a skill trajectory defined for a single grasp pose to any novel grasp. As a result, our method: (1) \textbf{assumes no prior object knowledge, such a 3D CAD model}, (2) \textbf{can operate with only RGB images if necessary}, (3) \textbf{is robust to noisy or partially missing depth data if using depth images}, and (4) \textbf{does not require any camera calibration}. 

Our method involves the robot collecting images (RGB or depth) of the grasped object by moving its EEF in front of a single external camera, in a self-supervised manner. As we will later show, this allows us to derive a framework that leverages these images to train a neural network that directly obtains the corrective transformation at skill deployment. As a result, with a few minutes of self-supervised data collection added to a standard pipeline that equips robots with skills, we can now adapt trajectories defined for a single grasp pose to different possible grasps.

We demonstrate the significance of our method through a series of real-world experiments, which quantify its accuracy and ability to adapt skill trajectories to novel grasp poses. Our experiments include adapting manually scripted trajectories for precise peg-in-hole insertion, as well as trajectories obtained with imitation learning for 7 everyday tasks, such as hammering in a nail, aligning a wrench with a nut, and inserting bread into a toaster. By evaluating 3 variants of our method and 5 depth-based pose estimation baselines we conclude that self-supervised RGB data results in the most accurate estimation of the corrective transformation. Specifically, we perform a total of 1360 real-world evaluations which show that, compared to the best-performing baseline, our method yields an average of \textbf{28.5$\%$} higher success rate when adapting skill trajectories to novel grasps.

\vspace{-.2ex}\section{Problem Formulation}\vspace{-.5ex}

\begin{figure*}[t!]
    \centering
    \includegraphics[width=\textwidth]{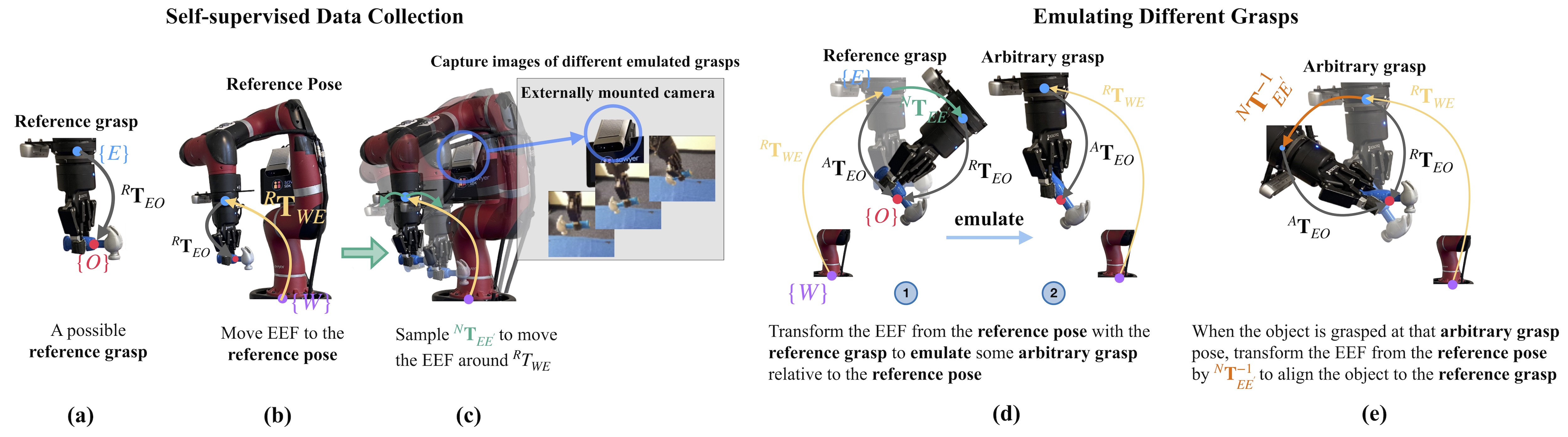}
    \vspace{-3ex}\small\caption{\textit{Self-supervised Data Collection: }(a) An example of a possible reference grasp. (b) The EEF at a potential reference pose $^{\textrm{R}}\mathbf{T}_{WE}$ in front of the external camera. The object is at the reference grasp. (c) With the object \textit{rigidly} grasped at the reference grasp, we sample and move the EEF to random poses ${^{\textrm{N}}}\mathbf{T}_{EE^{'}}$ relative to the reference pose to collect image-transformation pairs in a self-supervised manner that emulate different grasps. \textit{Emulating Different Grasps: }(d.1) By transforming the EEF and the object at the reference grasp by ${^{\textrm{N}}}\mathbf{T}_{EE^{'}}$, we emulate an arbitrary grasp with some grasp pose ${^{\textrm{A}}}\mathbf{T}_{EO}$ relative to the reference pose as it is shown in (d.2). (e) Then, if the object is grasped at that arbitrary grasp pose ${^{\textrm{A}}}\mathbf{T}_{EO}$ emulated by ${^{\textrm{N}}}\mathbf{T}_{EE^{'}}$, moving the EEF to ${^{\textrm{N}}}\mathbf{T}_{EE^{'}}^{-1}$ relative to the reference pose aligns the object to the reference grasp. }
    % \vspace{-4.7ex}
    \label{fig:grasp-emulate}
    \vspace{-3.8ex}
\end{figure*}

\label{sec:problem-formulation}
\textbf{Notations. }We define the following frames: $\{W\}$ which corresponds to the world frame, $\{E\}$ which corresponds to the end-effector's (EEF) frame and $\{O\}$ which is the frame of the grasped object. A transformation $\mathbf{T}_{WE}\in SE(3)$ defines the pose of frame $E$ relative to frame $W$. Also, we refer to the transformation $\mathbf{T}_{EO}$ as a \textbf{grasp pose}. We denote the different relative transformations between the same two frames using superscript notation. For example, ${{^\textrm{P}}}\mathbf{T}_{EO}$ and ${{^\textrm{Q}}\mathbf{T}_{EO}}$ denote two different grasp pose instantiations of the same object (for examples see Figure~\ref{fig:grasp-emulate}), ${{^\textrm{M}}}\mathbf{T}_{WE}$ and ${{^\textrm{K}}\mathbf{T}_{WE}}$ denote two different EEF poses expressed in the world frame, and so on. Additionally, we define the transformation $\mathbf{T}_{EE^{'}}$ expressed in frame $\{E\}$, to denote \textbf{displacement}, that is the EEF moves by $\mathbf{T}_{EE^{'}}$, from its current pose $\mathbf{T}_{WE}$ to the pose $\mathbf{T}_{WE}\mathbf{T}_{EE^{'}}$. We distinguish across different instantiations of EEF displacements using superscript notation, e.g. $^{\textrm{X}}\mathbf{T}_{EE^{'}}$. 
% Finally, we refer to the EEF transformation that \textbf{aligns} some grasp pose ${^\textrm{P}}\mathbf{T}_{EO}$ to some grasp pose ${^\textrm{Q}}\mathbf{T}_{EO}$ as the EEF displacement ${^\textrm{Z}}\mathbf{T}_{EE'}$ that moves the grasped object under ${^\textrm{P}}\mathbf{T}_{EO}$ to the same pose in the world frame as ${^\textrm{Q}}\mathbf{T}_{EO}$; that is, it satisfies: $\mathbf{T}_{WE} {^\textrm{Z}}\mathbf{T}_{EE'} {^\textrm{P}}\mathbf{T}_{EO} = \mathbf{T}_{WE}{^Q}\mathbf{T}_{EO}$ for any $\mathbf{T}_{WE}$. 
Finally, we refer to the EEF displacement that \textbf{aligns} an object grasped at pose ${^\textrm{P}}\mathbf{T}_{EO}$ with another grasp pose ${^\textrm{Q}}\mathbf{T}_{EO}$ as the transformation ${^\textrm{Z}}\mathbf{T}_{EE'}$, that moves the grasped object under ${^\textrm{P}}\mathbf{T}_{EO}$ to match the pose of ${^\textrm{Q}}\mathbf{T}_{EO}$ in the world frame; that is, it satisfies: $\mathbf{T}_{WE} {^\textrm{Z}}\mathbf{T}_{EE'} {^\textrm{P}}\mathbf{T}_{EO} = \mathbf{T}_{WE}{^Q}\mathbf{T}_{EO}$ for any $\mathbf{T}_{WE}$.

% Examples of grasp pose alignments can be seen in Figure~\ref{fig:corrective-transformation} (a) and (b). 

% Figure~\ref{fig:main-fig-with-frames} (a) and (b) show examples of grasp pose alignments.

\textbf{Motivation. }Consider a skill that enables the robot to manipulate an object (e.g. a hammer) grasped at a pose ${^{\textrm{S}}}\mathbf{T}_{EO}$ to complete some task (e.g., to hammer a nail). We refer to ${^{\textrm{S}}}\mathbf{T}_{EO}$ as the \textbf{skill grasp} and it denotes the grasp pose used to define the skill's trajectory (see Figure~\ref{fig:front-page} (a)). A skill's trajectory comprises a sequence of $H$ EEF displacements $\{^{t}\mathbf{T}_{EE^{'}}\}_{t=1}^H$ that make the grasped object track a sequence of $H$ poses $\{^{t}\mathbf{T}_{EE^{'}}{^{\textrm{S}}}\mathbf{T}_{EO}\}_{t=1}^H$ relative to some initial EEF pose $\mathbf{T}_{WE}$.  Examples of such trajectories tracked by a grasped hammer can be seen in Figure~\ref{fig:front-page}. 

% As an example, such a skill can be a vision-based neural network policy that receives images from an external camera and predicts EEF displacements.

% In line with \cite{johns2021coarse-to-fine,valassakis2022dome,dipalo2021learning}, we define a skill as either a trajectory of EEF poses the robot can track, or as a list of EEF velocities the robot can execute to complete a task.

%Given this, the pose of the object during skill acquisition in the robot's frame $\{W\}$ is denoted as
%\begin{equation*}
%    \mathbf{T}_{WO_S}=\mathbf{T}_{WE}{^{\textrm{S}}}\mathbf{T}_{EO}
%\end{equation*}
%for any EEF pose $\mathbf{T}_{WE}$. 
Now, consider a scenario during deployment of the skill, where the object is grasped differently to the skill grasp, at the grasp pose ${^\textrm{D}}\mathbf{T}_{EO}$, referred to as the \textbf{deployment grasp} (see Figure~\ref{fig:front-page} (b) and (c)). This can commonly occur due to the robot autonomously grasping the object (e.g., with GraspNet~\cite{Fang2020GraspNet1BillionAL}), due to external perturbations on the grasped object or due to a human handing the object to the robot.

Simply executing the skill's trajectory with the deployment grasp will likely result in task failure, as the object will follow a different trajectory to that of the skill grasp, as shown in Figure~\ref{fig:front-page} (b).  Specifically, as ${^\textrm{D}}\mathbf{T}_{EO} \neq {^{\textrm{S}}}\mathbf{T}_{EO}$, and hence the trajectory $\{^{t}\mathbf{T}_{EE^{'}}{^{\textrm{D}}}\mathbf{T}_{EO}\}_{t=1}^H \neq \{^{t}\mathbf{T}_{EE^{'}}{^{\textrm{S}}}\mathbf{T}_{EO}\}_{t=1}^H$ the skill will no longer be suitable to complete its designated task under the deployment grasp (Figure~\ref{fig:front-page} (b)). Motivated by this, we are interested in determining a \textbf{corrective transformation} ${^{\textrm{C}}}\mathbf{T}_{EE^{'}}$ that changes the EEF's trajectory during skill deployment such that the object follows the same trajectory it followed under the skill grasp (as shown in Figure~\ref{fig:front-page} (a) and (c)). That is ${^{\textrm{C}}}\mathbf{T}_{EE^{'}}$ aligns ${^{\textrm{D}}}\mathbf{T}_{EO}$ to ${^{\textrm{S}}}\mathbf{T}_{EO}$ for any EEF pose $\mathbf{T}_{WE}$:\begin{align}
\begin{split}
    \mathbf{T}_{WE}{^{\textrm{C}}}\mathbf{T}_{EE^{'}}{^\textrm{D}}\mathbf{T}_{EO}&=\mathbf{T}_{WE}{^{\textrm{S}}\mathbf{T}_{EO}} \Rightarrow\label{eq1}\\
{^{\textrm{C}}}\mathbf{T}_{EE^{'}}&={{^\textrm{S}}}\mathbf{T}_{EO}{^{\textrm{D}}}\mathbf{T}^{-1}_{EO}\,\,.
\end{split}
\end{align}
With the corrective transformation, we can adapt the skill's trajectory during skill deployment such that $ \{^{t}\mathbf{T}_{EE^{'}}{^{\textrm{C}}}\mathbf{T}_{EE^{'}}{^{\textrm{D}}}\mathbf{T}_{EO}\}_{t=1}^H = \{^{t}\mathbf{T}_{EE^{'}}{^{\textrm{S}}}\mathbf{T}_{EO}\}_{t=1}^H$. Figure~\ref{fig:front-page} (c) demonstrates the adapted sequence of poses followed by a hammer after applying the corrective transformation to the EEF's trajectory of Figure~\ref{fig:front-page} (b). 
% This setting assumes that for any deployment grasp the object is grasped such that the robot's gripper fingers do not obstruct the execution of the manipulation task. 
% Additionally, to adapt a skill in the presence of obstacles we assume that the skill learning method used is equipped with a collision-free path motion planner such as the ones proposed in~\cite{Karaman2011SamplingbasedAF}. 
We note that adapting a skill using the corrective transformation \textit{only} changes the EEF's trajectory; unlike regrasping methods, it does not change the grasp pose of the object in the robot's gripper.

\section{Method}\label{sec:method}
In this section, we present a method to determine the corrective transformation between \textit{any} pair of skill and deployment grasps of an object by leveraging images collected in a self-supervised manner in the real-world. This process requires no prior object knowledge or human time, no camera calibration, and it can work with either RGB or depth modalities or both. 

The process begins with a user handing the object to the robot's gripper at some arbitrary grasp pose which we refer to as the \textbf{reference grasp}. Then, the robot moves to different poses in front of an external camera in a self-supervised manner to emulate different possible grasps. By capturing images from the external camera and leveraging the robot's forward kinematics we train a vision-based alignment network that predicts an EEF displacement that can align any grasp to the reference grasp. As we later show, this allows us to obtain the corrective transformation between any pair of skill and deployment grasps by first aligning them to the reference grasp. This results in a task agnostic method where we can collect data for a grasped object \textit{once} and leverage that data to adapt skills across \textit{any} task for which the object is used and for \textit{any} skill or deployment grasps.

\subsection{Aligning grasps to a reference grasp}\label{sec:ref-grasp}

First, we define the \textbf{reference grasp} ${^\textrm{R}}\mathbf{T}_{EO}$. ${^\textrm{R}}\mathbf{T}_{EO}$ can be \textit{any} grasp pose, including that of the skill grasp. For generality, we assume that they are different, and in our experiments to define a reference grasp the user simply places the object in the EEF in a natural-looking pose for that object. Figure~\ref{fig:grasp-emulate} (a) shows an example of a possible reference grasp for a hammer. Note that in practice as we do not assume any prior object knowledge or access to a 3D CAD model we do not have access to the numerical value of the pose ${^\textrm{R}}\mathbf{T}_{EO}$.  Our initial goal is to determine how to align an object grasped at any possible grasp pose, i.e., any skill or deployment grasp, to the reference grasp, in the absence of any human intervention or prior object knowledge, and without requiring depth images or camera calibration, all of which are hard to achieve with existing pose estimation methods.

\begin{figure*}[t!]
    \centering
    \vspace{-.5ex}\includegraphics[width=\textwidth]{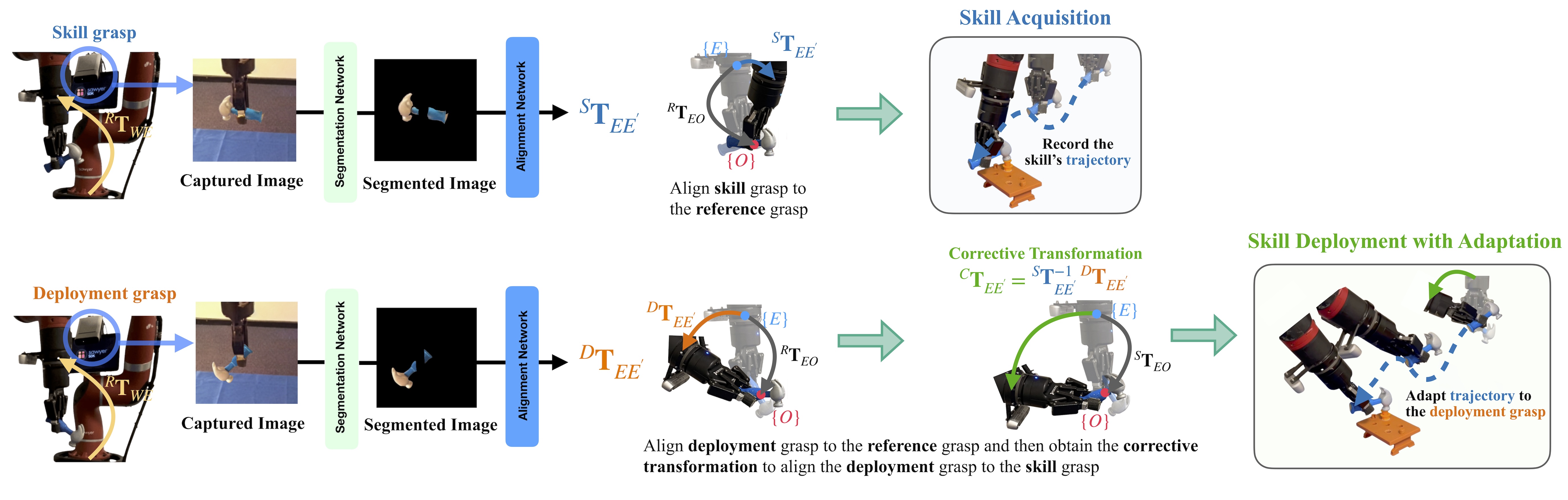}
    \caption{\textit{Top row:} Given a skill grasp, the EEF moves to the reference pose. Using our trained alignment network we obtain the transformation that aligns the skill grasp to the reference grasp. Then, a method suitable for equipping robots with skills is used to define the skill's trajectory. \textit{Bottom row:} During skill deployment for \textit{any} deployment grasp, we first obtain the transformation that aligns the deployment grasp to the reference grasp using our alignment network. This allows us to compute the corrective transformation which we use to adapt the skill's trajectory to the given deployment grasp. } 
    \vspace{-3ex}
    \label{fig:method-deployment}
\end{figure*}

\textbf{Self-supervised data collection. }Towards this end, we seek a way to \textit{emulate} the appearance of different possible grasps in the robot's EEF autonomously, in a self-supervised manner. To achieve this, we begin by moving the EEF to a \textbf{reference pose} $^{{\textrm{R}}}\mathbf{T}_{WE}$ with the object grasped at the reference grasp, as shown in Figure~\ref{fig:grasp-emulate} (b). The reference pose can be arbitrary as long as the grasped object is clearly visible to the camera. Then, from $^{{\textrm{R}}}\mathbf{T}_{WE}$, we sample and move the robot to random poses ${^{\textrm{N}}}\mathbf{T}_{EE^{'}}$ relative to the reference pose. Throughout this process we do not manually move the object in the gripper, instead, the object remains \textit{rigidly} grasped at the reference grasp, requiring no human intervention. At every EEF pose $^{{\textrm{R}}}\mathbf{T}_{WE}{^{\textrm{N}}}\mathbf{T}_{EE^{'}}$ we capture an image of the grasped object, $I$, and record the inverse transformation ${{^\textrm{N}}}\mathbf{T}^{-1}_{EE^{'}}$, as can be seen in Figure~\ref{fig:grasp-emulate} (c) and Figure~\ref{fig:data-col} in the appendix. This way we record a dataset of $M$ image-transformation pairs $\mathcal{D}:=\{(I, {{^\textrm{N}}}\mathbf{T}^{-1}_{EE^{'}})_i\}_{i=1}^M$ in a self-supervised manner. 

Every random EEF displacement ${^{\textrm{N}}}\mathbf{T}_{EE^{'}}$ emulates a different, arbitrary grasp with some grasp pose ${{^\textrm{A}}}\mathbf{T}_{EO}$ (whose numerical value is unknown) \textit{relative to the reference pose}, as shown in Figure~\ref{fig:grasp-emulate} (d.1) and (d.2). As a result, every sample $(I, {{^\textrm{N}}}\mathbf{T}^{-1}_{EE^{'}})_i$ in our dataset $\mathcal{D}$ contains an image $I$ that depicts how the grasped object would appear to the camera if the object was grasped at that arbitrary grasp pose ${{^\textrm{A}}}\mathbf{T}_{EO}$ \textit{and the EEF was at the reference pose}, as shown in Figure~\ref{fig:grasp-emulate} (d.2). And every ${{^\textrm{N}}}\mathbf{T}^{-1}_{EE^{'}}$ corresponds to the transformation that we need to apply to the robot's EEF at the reference pose to align the object at that arbitrary grasp pose ${{^\textrm{A}}}\mathbf{T}_{EO}$ to the reference grasp, as shown in Figure~\ref{fig:grasp-emulate} (e). This is true since $^{{\textrm{R}}}\mathbf{T}_{WE}{^{\textrm{N}}}\mathbf{T}_{EE^{'}}{{^\textrm{R}}}\mathbf{T}_{EO}={^{{\textrm{R}}}\mathbf{T}_{WE}}{{^\textrm{A}}}\mathbf{T}_{EO}$ and as a result $^{{\textrm{R}}}\mathbf{T}_{WE}{{^\textrm{N}}}\mathbf{T}^{-1}_{EE^{'}}{{^\textrm{A}}}\mathbf{T}_{EO}={^{{\textrm{R}}}}\mathbf{T}_{WE}{{^\textrm{R}}}\mathbf{T}_{EO}$. 

For pseudocode detailing our data collection procedure please see Algorithm~\ref{algo:train} in the appendix.

\subsection{Alignment Network}\label{sec:alignment-network}
After collecting our dataset $\mathcal{D}$ we train an \textbf{alignment network} which is a function parameterised by $\theta$, $f_\theta: \mathbb{R}^{H\times W\times C}\rightarrow SE(3)$ using supervised learning to predict poses ${{^\textrm{N}}}\mathbf{T}^{-1}_{EE^{'}}$ given camera images ($H$: height, $W$: width of the image and $C=3$ for RGB and $C=1$ for depth).

In practice, when the robot grasps the object at a pose different from the reference grasp, the gripper's fingers will occlude parts of the object differently to the occlusions captured in images in $\mathcal{D}$. Hence, we need to ensure that the alignment network's predictions are robust to any occlusion caused by the gripper when grasping the object. To achieve this, before training $f_\theta$, we segment the EEF and background from each image $I$ in $\mathcal{D}$ using a pre-trained optical flow network \cite{xu2022unifying} which we deploy similarly to \cite{Boerdijk2020SelfSupervisedOS} and perform additional data augmentations. This way, we ensure that our alignment network's predictions are \textit{object-centric} and robust to any object occlusions caused by the gripper or previously unseen image variations. For more implementation details we refer the reader to the appendix I.\ref{sec:dataset-segmentation}-I.\ref{sec:app-training}.

At test time, given a grasp, to obtain the transformation ${{^\textrm{N}}}\mathbf{T}^{-1}_{EE^{'}}$, we first move the EEF to the reference pose $^{\textrm{R}}\mathbf{T}_{WE}$. Then, we capture a live image and segment everything but the grasped object. The segmented image is then passed through $f_\theta$ to obtain ${{^\textrm{N}}}\mathbf{T}^{-1}_{EE^{'}}$. In practice, we obtain ${{^\textrm{N}}}\mathbf{T}^{-1}_{EE^{'}}$ in a visual servoing (VS) manner. During the VS process, we leverage our robot's redundant DOFs and motion-planning to ensure that we avoid self-collisions and joint limits. We refer the reader to the appendix I.\ref{sec:visual-servoing} for a derivation of the VS process using our alignment network's predictions. Finally, as the external camera is rigidly mounted to the robot during data collection and deployment of $f_\theta$, our method is also \textit{independent} of camera calibration.

%  \begin{figure}[t!]
%     \centering
%     \vspace{-0ex}\includegraphics[width=.43\textwidth]{figs/alternative_method_grasp_alignments.pdf}
%     \caption{(a) Examples of a possible skill grasp, reference grasp and deployment grasp. (b) After aligning the skill grasp and the deployment grasp to the reference grasp (c) we obtain ${^\textrm{S}}\mathbf{T}_{EO}$ and ${^\textrm{D}}\mathbf{T}_{EO}$ respectively, which use to calculate the corrective transformation ${^{\textrm{C}}}\mathbf{T}_{EE^{'}}={{^\textrm{S}}}\mathbf{T}_{EE^{'}}^{-1}{^{\textrm{D}}}\mathbf{T}_{EE^{'}}$.}\vspace{-3.5ex}
%     \label{fig:corrective-transformation}
% \end{figure}
\begin{figure*}[t!]
\captionsetup{labelfont={color=black}}

    \centering
    \includegraphics[width=\textwidth]{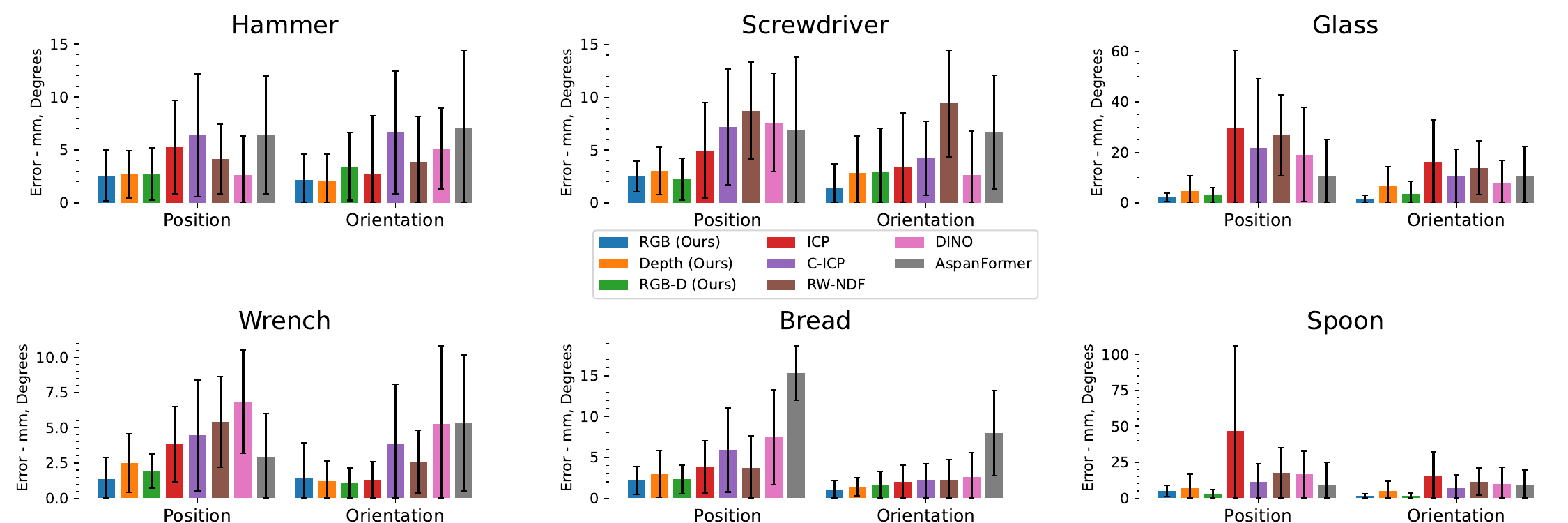}
\caption{Mean and standard deviation error in computing the corrective transformation between grasps averaged across all DoFs for the position and orientation for the 6 everyday objects ({lower is better}).}\vspace{-3.5ex}
    \label{fig:accuracy-main-results}
\end{figure*}

\vspace{-.5ex}\subsection{Corrective Transformation}\label{sec:corr-trans}\vspace{-.5ex}

Now that we have a way to align any grasp to the reference grasp, consider a skill grasp ${^{\textrm{S}}}\mathbf{T}_{EO}$. First given the skill grasp we deploy our alignment network $f_\theta$ to obtain the transformation that aligns the skill grasp to the reference grasp. For clarity, we denote that transformation as ${^{\textrm{S}}}\mathbf{T}_{EE^{'}}$ (instead of ${^{\textrm{N}}}\mathbf{T}^{-1}_{EE^{'}}$). Then, a method suitable for equipping robots with skills is used to define the skill's trajectory (e.g., an imitation learning method like \cite{valassakis2022dome, johns2021coarse-to-fine}), as shown in the top row of Figure~\ref{fig:method-deployment}.  Then, at skill deployment, given \textit{any} novel deployment grasp, we obtain the transformation that aligns that grasp to the reference grasp in an identical manner using the alignment network. For clarity, we denote that transformation as ${^{\textrm{D}}}\mathbf{T}_{EE^{'}}$. Given ${^{\textrm{S}}}\mathbf{T}_{EE^{'}}$ and ${^{\textrm{D}}}\mathbf{T}_{EE^{'}}$, we can calculate the corrective transformation ${^{\textrm{C}}}\mathbf{T}_{EE^{'}}$ trivially, that is ${^{\textrm{C}}}\mathbf{T}_{EE^{'}}={{^\textrm{S}}}\mathbf{T}_{EE^{'}}^{-1}{^{\textrm{D}}}\mathbf{T}_{EE^{'}}$. Then, given the corrective transformation, we can immediately adapt a skill's trajectory, as shown in the bottom row of Figure~\ref{fig:method-deployment} and discussed in Section \ref{sec:problem-formulation}.

To see why we can obtain the corrective transformation this way, note that ${^{\textrm{S}}}\mathbf{T}_{EE^{'}}$ and ${^{\textrm{D}}}\mathbf{T}_{EE^{'}}$ align the skill grasp ${^\textrm{S}}\mathbf{T}_{EO}$ and deployment grasp ${^\textrm{D}}\mathbf{T}_{EO}$ to the reference grasp ${^\textrm{R}}\mathbf{T}_{EO}$ respectively, that is: ${^{\textrm{S}}}\mathbf{T}_{EE^{'}}{^\textrm{S}}\mathbf{T}_{EO}={{^\textrm{R}}\mathbf{T}_{EO}}={^{\textrm{D}}}\mathbf{T}_{EE^{'}}{^\textrm{D}}\mathbf{T}_{EO}$. Hence, it follows that: ${^\textrm{S}}\mathbf{T}_{EO}{^\textrm{D}}\mathbf{T}^{-1}_{EO}={^{\textrm{S}}}\mathbf{T}^{-1}_{EE^{'}}{^{\textrm{D}}}\mathbf{T}_{EE^{'}}$ and as a result from Equation~\ref{eq1} we can see that ${^{\textrm{C}}}\mathbf{T}_{EE^{'}}={{^\textrm{S}}}\mathbf{T}_{EE^{'}}^{-1}{^{\textrm{D}}}\mathbf{T}_{EE^{'}}$. Note that the corrective transformation is not dependent to the reference pose or any EEF pose relative to the world frame. This allows us to adapt the skill's trajectory across the whole task space of the robot.

For pseudocode detailing the deployment of our method please see Algorithm~\ref{algo:skill-dep} in the appendix.

Finally, we note that the alignment network is agnostic to the skill's trajectory and its designated task. This is very important: once the alignment network is trained for a particular grasped object, we can re-use that same alignment network for \textit{any} task with that same object, without further training. A video demonstrating this ability can be found at the bottom of our webpage at \href{https://www.robot-learning.uk/adapting-skills}{www.robot-learning.uk/adapting-skills}.

\vspace{-.7ex}\section{Experiments}\label{sec:experiments}\vspace{-.7ex}

We evaluate our method by performing three sets of real-world experiments totaling 1360 real-world evaluations. Through these experiments, we answer the following questions: 1) How accurate are the corrective transformations obtained by our method? 2) Can our method adapt skills to novel deployment grasps for (a) precise tasks and (b) tasks learned with imitation learning? 3) What is the best modality to determine the corrective transformation in the real-world; RGB, depth or both combined (RGB-D)? Videos of our real-world experiments can be found on our webpage at \href{https://www.robot-learning.uk/adapting-skills}{www.robot-learning.uk/adapting-skills}.
% Videos of our experiments can be found at our supplementary material and on our anonymous website:  \url{https://sites.google.com/view/self-supervised-adapt-skills}

% To this end, we design the following experiments. (1) To answer question 1, in section~\ref{sec:experiments1}  we benchmark and quantify our method's accuracy in determining the corrective transformation on objects used for six everyday tasks (see Figure~\ref{fig:everyday-objects}) when using different input modalities. %Additionally, in this experiment we compare our method against ICP \cite{ICP}, Colour-ICP \cite{park2017colored} and a real-world NDF \cite{simeonovdu2021ndf} variant which we implement,
% (2) To answer question 2 (a), in Section~\ref{sec:peg-in-hole-evaluation} we quantify our method's ability to accurately adapt a precise peg-in-hole insertion skill to different novel deployment grasps (see Figure~\ref{fig:peg-in-hole-task}). (3) To answer question 2 (b), in Section~\ref{sec:dome-evaluation} we evaluate our method's ability to adapt six everyday skills learned via imitation learning to novel deployment grasps (see Figure~\ref{fig:everyday-objects}). For all experiments, we train and evaluate 3 variants of our method: one based only on RGB, one based on depth, and one based on the RGB-D modality. Finally, drawing from the results obtained in the experiments we answer question 3.

\textbf{Implementation details. }For our experiments, we use a 7 DoF Rethink Sawyer Robot to which we mount a Microsoft Azure Kinect camera. To determine the reference pose $^\textrm{R}\mathbf{T}_{WE}$, we empirically evaluate the quality of depth images for the objects used for evaluation at various EEF poses and select the best one. We found this to be crucial to obtain good performance for the baselines, as the quality of the depth images varied significantly based on an object's pose, unlike our method that is robust to this. We allow an average of approximately $5$ minutes of real-world data collection, during which we sample random poses around $^\textrm{R}\mathbf{T}_{WE}$ in the range of $30\textrm{cm}$ for each of the DoFs relating to position, and $60^\circ$ for each of the DoFs relating to orientation. This range is not limiting and can be trivially adjusted to any value to accommodate any task requirements; we found that this range was sufficient for our tasks. Preprocessing the collected data and training takes approximately $15$ minutes on an NVIDIA GeForce RTX 3080 Ti. 
% This way, we obtain the best possible performance both for the depth-based variants of our method and the baselines we compare against. To collect our dataset $\mathcal{D}$, we sample random poses around $^\textrm{R}\mathbf{T}_{WE}$ in the range of $30\textrm{cm}$ for each of the DoFs relating to position, and $60^\circ$ for each of the DoFs relating to orientation. We collect our dataset by allowing $20$ minutes of real-world data collection, during which we store RGB-D observations to train all variants of our method. Our camera returns images of $1280\times 720$ resolution, which we crop and resize to $128\times 128$. 
% To obtain a segmentation network for each grasped object we use a pre-trained flow network \cite{xu2022unifying}, which we deploy on our collected dataset in a similar manner to \cite{Boerdijk2020SelfSupervisedOS}. 
We refer the reader to appendix I.\ref{sec:app-training}, I.\ref{sec:net-arch} and I.\ref{sec:data-col-time} for further implementation details, a description of our network architectures and a discussion on the effect of data collection time on our method's performance.

\vspace{-1ex}\subsection{How accurate are the corrective transformations obtained by our method?}\label{sec:exp-accuracy}\label{sec:experiments1}

% \subsubsection{Motivation} We design the following experiment to answer questions 1 and 2 set out at the beginning of our experiments. Our objective is to evaluate the accuracy of our method in determining the corrective transformation between object grasps when using different input modalities. Hence, in this experiment, we do not evaluate our method's ability to adapt a skill and no skill learning is involved. 
In this experiment, we quantify the accuracy of our method in determining the corrective transformation between pairs of object grasps. To perform our evaluation, we use 6 objects for the everyday tasks shown in Figure~\ref{fig:everyday-objects}. That is, a plastic \textbf{hammer}, a plastic \textbf{screwdriver}, a plastic \textbf{bread}, a metallic \textbf{spoon}, a plastic \textbf{wrench}, and a semi-transparent wine \textbf{glass}.
% That is we first move the robot to the reference pose and hand the object to the EEF at a random pose. This defines a potential skill grasp. We then deploy our method to compute ${^{S}\mathbf{T}_{EE'}}$. Without changing the grasp, we move the EEF to a random pose to emulate a possible deployment grasp. Similarly to the skill grasp, we obtain  ${^{D}\mathbf{T}_{EE'}}$ and compute the corrective transformation. If the corrective transformation is accurate, the EEF should return to the reference to align the deployment grasp to the skill grasp. By computing the error to the reference pose using the robot's forward kinematics we can  quantify the error in the obtained corrective transformation.

\textbf{Evaluation procedure. }As we do not have access to the true pose of the objects, we evaluate our method using the robot's forward kinematics. We refer the reader to appendix II.\ref{sec:acc-eval-proc} for a detailed description of our evaluation procedure. For each of the six objects, we randomly set 4 skill grasps and for each skill grasp, we evaluate the corrective transformation for 5 random  deployment grasps leading to 20 evaluations per object. For each evaluation, we allow our method 5 seconds of visual servoing to align each grasp to the reference grasp. 

\textbf{Baselines. }As we have no information about the objects' CAD models, we compare all variants of our method (RGB, Depth, RGB-D) on the same objects against 5 baselines that can be deployed without requiring any prior object knowledge: 1) ICP, 2) Color-ICP (C-ICP), 3) a real-world variant of neural descriptor fields~\cite{simeonovdu2021ndf} (RW-NDF) that we implement,  and two correspondence based methods that leverage 4) DINO ViT (DINO) \cite{Caron2021EmergingPI,amir2021deep} and 5) AspanFormer\cite{chen2022aspanformer}. 

For ICP and C-ICP we capture a segmented depth image of the grasped object under the skill and deployment grasps and compute the relative pose between them in the camera's frame. Then, by leveraging the camera's extrinsics we obtain the corrective transformation. Additionally, we note that we tried scanning the object by moving it in front of the external camera to obtain a more complete 3D object model before using ICP and C-ICP, but performance did not improve and was almost identical. For DINO and AspanFormer we first obtain correspondences between the RGB images of the skill and deployment grasps and obtain their relative pose by leveraging the corresponding depth images and singular value decomposition~\cite{Arun1987LeastSquaresFO}.    Identically to our method's evaluation, we allow each baseline 5 seconds of visual servoing. We refer the reader to the appendix II.\ref{sec:baselines} for more details on the baselines implementations.
\textbf{Results. }Figure~\ref{fig:accuracy-main-results} shows the mean and standard deviation error of the corrective transformation averaged separately across the 20 evaluations for the DoFs relating to position and orientation for each object and method. The quantitative results of Figure~\ref{fig:accuracy-main-results} can be seen in the appendix Table~\ref{table:reorient-accuracy}. As shown, the RGB, Depth and RGB-D variants of our method perform better, on average, than the 5 baselines both with respect to position and orientation error. Specifically, the best-performing variant overall from our methods is RGB with a 2.65mm mean position error and 1.50$^\circ$ mean orientation error. In addition to the high accuracy, the results of Figure~\ref{fig:accuracy-main-results} also confirm that our method is robust to object occlusions caused by the gripper's fingers under different grasps. From the baselines, AspanFormer obtains the lowest mean error in position (8.59mm) and DINO for orientation (5.56$^\circ$). Hence, RGB obtains a 69.1$\%$ increase in position accuracy compared to AspanFormer and a 73.0$\%$ increase in orientation accuracy compared to DINO, averaging {at least} a \textbf{71.1}$\%$ increase in accuracy when compared to all the baselines.
% Interestingly, we observe that ICP performs significantly better than the results reported for ICP in \cite{Devgon2020OrientingN3}.

The performance difference is most profound for the spoon and glass objects (see Figure~\ref{fig:accuracy-spoon-glass} in the appendix), where all the baselines struggle to accurately determine the corrective transformation. We attribute this difference to the missing depth data for these objects which is due to their textures; that is the spoon is metallic and the glass semi-transparent. As shown in the appendix Figures~\ref{fig:spoon-pcd} and~\ref{Fig:glass-pcd} the point clouds for these objects have very low quality. On the other hand, our depth-based variant shows robustness to missing depth data as it was trained \textit{directly} on the real-world depth images for each object. Nevertheless, it is still less accurate when compared to the RGB and RGB-D variants.
\subsection{Can our method adapt skills to novel deployment grasps for precise tasks?}\label{sec:peg-in-hole-evaluation}
% \begin{wrapfigure}{r}{.5\textwidth}
%     \centering
%     \vspace{-.39cm}\includegraphics[width=0.4\textwidth]{figs/peg-in-hole.pdf}\vspace{-.27cm}
%     \caption{The peg and the base with 4 different holes, each with a different insertion tolerance used for the peg-in-hole task.}\vspace{-.56cm}
%     \label{fig:peg-in-hole-task}
% \end{wrapfigure}
In this experiment, we evaluate the ability of our method to adapt skill trajectories for precise tasks. To this end, we 3D printed a base with 4 different holes and a peg. The 4 holes have insertion tolerances of $\{2, 4, 8, 12\}$mm on each side of the peg, as shown in Figure~\ref{fig:everyday-objects}. In this setting, we assume that the pose of the base is known. This setup may correspond to an industrial assembly setting where the pose of the insertion hole is known, but there is uncertainty on the grasp pose of the object (e.g. because the robot autonomously grasps it).  In this setting, manually programming a separate trajectory for every possible grasp is infeasible. Instead, it is significantly more efficient to design a single insertion skill for some skill grasp and deploy our method to adapt the insertion trajectory for each deployment grasp.

\textbf{Evaluation procedure. }First, for each hole, we hand the peg to the robot to define a skill grasp and we manually program a trajectory of poses to insert the peg in each of the 4 holes which we track with a position controller. 
% This allows us to evaluate our method in isolation, independently of errors that may be due to a skill being a policy learned from data using methods such as imitation or reinforcement learning. 
Then, we randomly change the pose of the peg in the gripper and consider that as a deployment grasp. Given the deployment grasp, we deploy our method to adapt the insertion trajectory as discussed in section~\ref{sec:problem-formulation}. We evaluate 5 different deployment grasps for each hole totalling 20 evaluations for each method.

\textbf{Baselines. }Based on the results of Section~\ref{sec:experiments1}, we compare our method against the best non-learning-based baseline, ICP, and the best learning-based baseline, DINO. As we are interested in success rate, we selected ICP because it outperforms C-ICP in 4 out of the 6 objects, although its average error is higher as it is negatively affected by the "Glass" and "Spoon" objects. Similarly, we selected DINO as it outperforms RW-NDF and AspanFormer in more objects when accounting both for the translation and orientation errors. We evaluate the baselines identically to our method.

\textbf{Results. } Table~\ref{table:peh-in-success-rate} shows the success rate results for all methods. All variants of our method, outperform on average both ICP and DINO. Both ICP and DINO successfully adapt most deployment grasps for the holes with tolerances 8mm and 12mm, but fail to adapt the majority of the deployment grasps for the 2mm and 4mm holes. As shown, the RGB variant of our method performs best overall, outperforming ICP and DINO on average by \textbf{25}$\%$ and \textbf{30}$\%$ respectively, only failing to adapt one deployment grasp for the hole with the lowest tolerance of 2mm. The RGB-D variant also performs well but fails to adapt 3 deployment grasps for the 2mm tolerance hole. On the other hand, the Depth variant has the lowest performance across our method's variants.
% it fails to adapt 1 skill even for the 8mm tolerance hole. 

\input{peg_hole_table} The results suggest that our method can accurately determine the corrective transformation between grasps to adapt skill trajectories even for tasks with low error tolerance. However, our results also indicate that the RGB modality is crucial for high precision. Interestingly, when combining RGB with depth, the performance seems to be better than using only depth but lower compared to using only RGB. This is likely because, during training, our alignment network is optimized to give a certain weight to RGB and a certain weight to the depth modality based on the depth values recorded in the training dataset. However, if the depth images observed during testing do not match those observed during training, likely due to the high level of noise in the depth signal, this can negatively affect the alignment network's predictions.

% • \textbf{PARAGRAPH 1.} In this paragraph we will give a detailed explanation of the peg-in-hole accuracy tasks for the 4 different holes with different insertion tolerances. Here we will also include a figure of the holes and peg.

% • \textbf{PARAGRAPH 2.} In this paragraph we will present the results in a table like the one shown in Table~\ref{table:peh-in-hole-accuracy}. We may include an evaluation paragraph for each method as suggested in the previous subsection.

\subsection{\hspace{-.1cm}Can our method adapt skills learned with imitation learning to novel deployment grasps?}\label{sec:dome-evaluation}

\begin{figure}[t!]
    \centering
    \vspace{-0cm}\includegraphics[width=.45\textwidth]{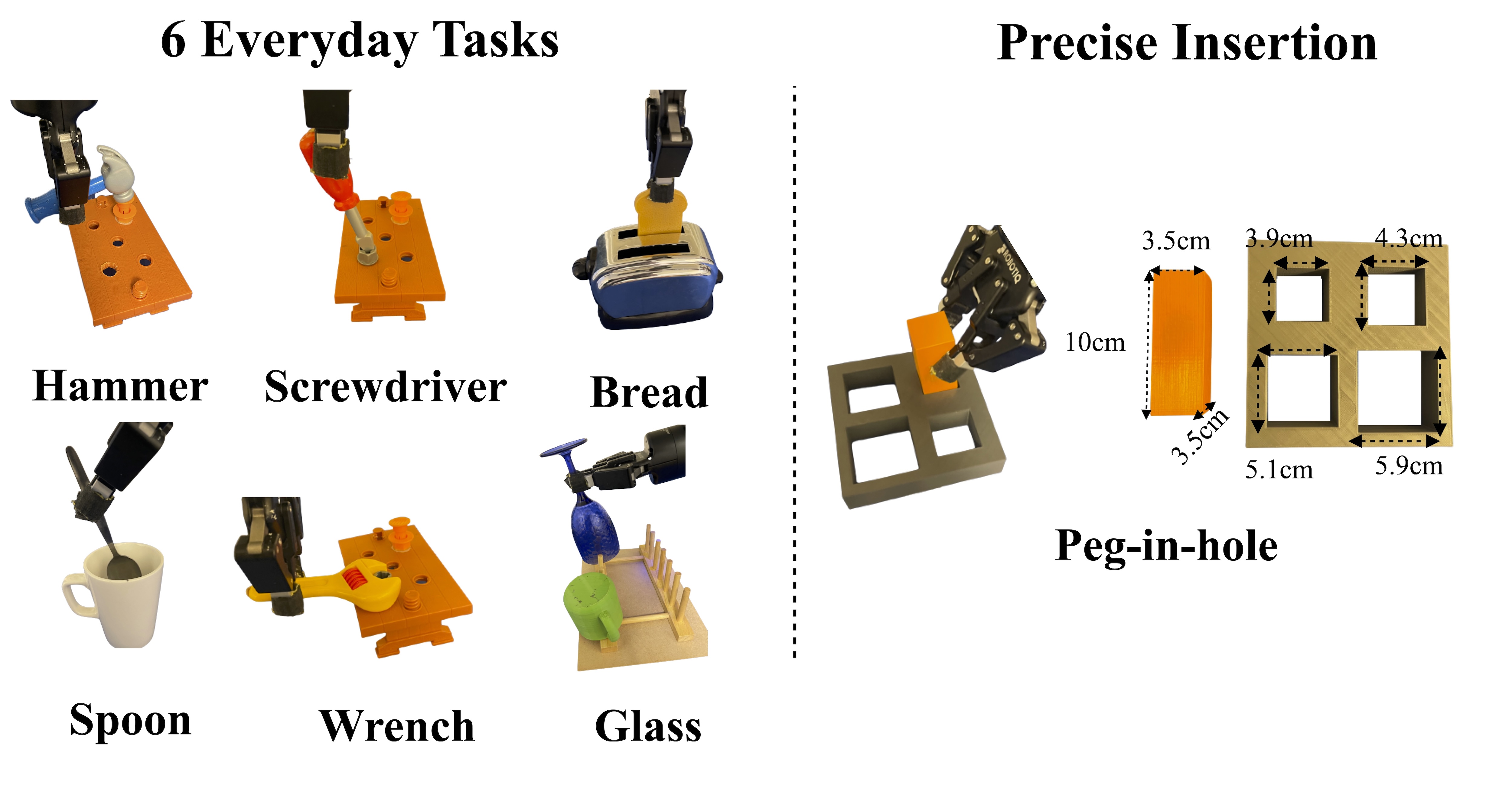}\vspace{-.4cm}
    \caption{The 6 everyday objects and tasks and the peg-in-hole task used in the experiments.}\vspace{-3ex}
    \label{fig:everyday-objects}
\end{figure}

In this experiment, we are interested in evaluating our method as a modular component added to a skill acquisition pipeline where skill trajectories are obtained using imitation learning. This setup corresponds to a realistic robot learning setup where a user may teach a robot skills using human demonstrations. Specifically, we use the one-shot imitation learning method DOME \cite{valassakis2022dome} to equip our robot with skills to solve the 6 everyday tasks shown in Figure~\ref{fig:everyday-objects}. DOME is a vision-based imitation learning method whose action space comprises EEF twists. We provide a brief overview of DOME in appendix II.\ref{sec:dome} and refer the reader to \cite{valassakis2022dome} for more details.

\textbf{Evaluation procedure.} We teach the robot skills using DOME to solve the following six everyday tasks: A \textit{Hammer} task requiring the robot to knock a plastic nail using a plastic hammer into a receptacle. A \textit{Screwdriver} task requiring the robot to fully insert the tip of the plastic screwdriver into the slit of a screw. This task has very low tolerance, requiring millimetre precision. A \textit{Bread} task requiring the robot to insert a plastic bread into the slit of a toaster. A \textit{Spoon} task requiring the robot to insert the spoon into a mug and stir. A \textit{Wrench} task that requires the robot to insert a plastic nut into the wrench's head. Finally, a \textit{Glass} task where the robot needs to place a wine glass standing upright on a wooden rack. We chose these tasks to represent a variety of challenges and tolerances ranging from several cm of tolerance (\textit{Spoon}) to around 1 mm of tolerance (\textit{Screwdriver}).
For each task, first, we hand the relevant object to the robot's EEF to define a skill grasp and provide a demonstration using DOME. Then, we manually sample novel deployment grasps approximately within $10\text{cm}$ and $90^\circ$ of the skill grasp for all the axes not constrained by the gripper's fingers and deploy DOME along with our method to adapt each skill's twists as described in section \ref{sec:problem-formulation}.  We perform this process for 10 different grasps for each task and record the success rate of each skill adaptation trial, totaling 60 evaluations per method. For each evaluation, we also randomize the task space configuration and consider a trial successful if the task is completed as in the demonstration. Finally, as in section~\ref{sec:peg-in-hole-evaluation}, we compare our method against the ICP and DINO baselines.

\textbf{Results. }Table \ref{table:dome-results} shows the skill adaptation success rate for all the methods. All variants of our method outperform the baselines in the majority of trajectory adaptation trials, apart from the Screwdriver task where ICP outperforms both the Depth and RGB-D variants and the Bread task where DINO is the best-performing method along with RGB-D. Overall, the best-performing variant of our methods is RGB with an average success rate of 75 $\%$, outperforming on average ICP and DINO by \textbf{32}$\%$ and \textbf{37}$\%$ respectively. We observe that all variants of our method can successfully adapt trajectories obtained by DOME to most deployment grasps, especially for the Hammer, Spoon and Glass tasks. The depth and RGB-D variants failed twice to adapt the trajectory for the Glass task. These failures likely relate to the semi-transparent texture of the glass which yields poor depth quality; an observation also made in section~\ref{sec:experiments1}. 

% Additionally, even small errors in determining the corrective transformation for the Glass task can compound to large deviations from the demonstrated interaction trajectory. This is because the interaction trajectory to place the glass on the rack was long. Hence, a small deviation at the beginning of the trajectory can compound to a large error relative to the pose at which the glass is placed on the rack. This is amplified when accounting for errors caused both due to our method and DOME.
\input{dome_results.tex}All variants of our method fail almost completely to adapt the trajectory obtained by DOME for the Screwdriver task. We attribute this failure mainly to DOME's error in accurately approaching the screw as our method yielded small inaccuracies in determining the corrective transformation in the order shown in Figure~\ref{fig:accuracy-main-results}. Although small, the errors from both methods can result in failure when accumulated, especially for a high-precision task like the Screwdriver task. Similar reasoning applies to the Bread task, which, however, has a higher tolerance to error compared to the Screwdriver and, subsequently, a higher success rate.

Following these results and the results obtained in Section~\ref{sec:peg-in-hole-evaluation} we can see that the RGB variant yields on average a \textbf{28.5}$\%$ higher success rate compared to ICP and \textbf{33.5}$\%$ when compared to DINO.

\subsection{What is the best modality to determine the corrective transformation between pairs of object grasps in the real-world, RGB, depth or both combined (RGB-D)?} In our experiments, we observed that all modalities perform similarly when trained using our method. Overall, RGB is crucial when dealing with objects for which the depth quality is poor, especially when compared to depth-based registration methods as demonstrated in the previous sections. Further, even for the objects for which the depth quality was high, the performance was on par with that of RGB. In fact, we observed that RGB performed better for the peg-in-hole task. Hence, we can conclude that the RGB modality is the most robust overall. It performs well for all types of objects, causing no performance degradation compared to depth on all the experiments we conducted.

% \textit{Can we use the proposed method to adapt skills to novel deployment grasps for precise and everyday tasks learned with imitation learning?} Following the evaluations performed in Section \ref{sec:peg-in-hole-evaluation} and Section~\ref{sec:dome-evaluation}, we observed that our method could successfully adapt skills to the majority of the deployment grasps evaluated both for inserting a peg into holes of different tolerances and six everyday tasks demonstrated to the robot using the one-shot imitation learning method DOME.
% • \textbf{PARAGRAPH 1.} In this paragraph we will give a detailed explanation of the skill adaptation task for the one-shot imitation learning method DOME and why we picked that method for evaluation, that is because it is easy and quick essentially.

% • \textbf{PARAGRAPH 2.} Provide a brief overview of how DOME works and how we use our method to adapt skills using DOME.

% • \textbf{PARAGRAPH 3.} Present our success rate results in a table format as shown in Table~\ref{table:dome-results}.

\section{Discussion}\label{sec:discussion}

\subsection{Limitations}

We now highlight some important limitations of our method. Firstly, to obtain an alignment network for an object our method needs to collect data for a few minutes. As such, for basic low-tolerance tasks, simple baselines such as ICP may be preferred as they do not require data collection. However, for real-world tasks requiring precision, our method is \textit{necessary} to overcome errors from calibration or missing depth data. And, as our alignment network can be reused across \textit{any} task for an object without further training, if we aim to adapt skills across multiple grasps and tasks the required data collection time becomes comparably negligible.

Secondly, our method assumes that the object under the deployment grasp is grasped such that the gripper's fingers do not obstruct task execution. To address this assumption future work can trivially couple our method with affordance-based grasping approaches, e.g., \cite{Hadjivelichkov2022affcorrs} or imitation learning approaches like DOME \cite{valassakis2022dome} that can show to the robot which parts of an object to grasp. Additionally, this assumption is not practically limiting as most deployment grasps can be adapted, and as we showed in our experiments we can adapt skill trajectories over a wide range of grasp poses. Also, future work could investigate extending our method to detect such scenarios and perform regrasping by leveraging the corrective transformation. 

Thirdly, our method assumes that for a given deployment grasp executing the adapted trajectory results in the same kinodynamic interaction with the environment as with the skill grasp. In most scenarios, this can be achieved by leveraging our robot's redundant degrees of freedom and motion planning as we do in our experiments. However, for grasps where the adapted trajectory cannot complete a task due to issues relating to the robot's kinematics, future work can couple our method with approaches that determine the initial robot configuration such that skill execution succeeds\cite{vosylius2022where}. 

% Fourthly, our method is not explicitly designed to handle \textbf{symmetric objects}. This may lead our network to make inaccurate predictions around an object’s axis of symmetry, as there may be more than one corrective transformation that aligns two grasps. 
Finally, compared to methods such as \cite{florence2018dense, simeonovdu2021ndf, amir2021deep, pmlr-v205-goodwin23a}, in the current setup {same-category generalization} is not possible with our approach. However, these methods assume prior access to category-specific training data which may not be readily available, and rely on depth images which hinder the performance as shown in our experiments. Instead, our method addresses these issues. In future work, we aim to study whether training our method directly on image descriptors extracted from our dataset using pertained vision models, such as DINO ViTs \cite{Caron2021EmergingPI}, allows our method to generalize to novel objects of the same category.

\subsection{Conclusions }In this work, we proposed an autonomous, self-supervised method that enables the adaptation of skill trajectories defined for a single object grasp pose to any novel grasp pose at skill deployment, without any prior knowledge about the grasped object. Through multiple real-world experiments, we show that our method enables skills acquired through imitation learning, for several everyday tasks, to be adapted to different grasps at deployment without the robot needing to re-learn or fine-tune the skill itself. Importantly, our results demonstrate that RGB data collected in a self-supervised manner is the best modality that obtains the highest skill adaptation performance when compared to depth-based alternatives including several state-of-the-art pose estimation methods.
% \bibliographystyle{ieeetr}
% \bibliography{references}

% \vspace{-1ex}\scriptsize\printbibliography[title={References}]\normalsize
\bibliographystyle{ieeetr}
\vspace{-1ex}\footnotesize\bibliography{arxiv}\normalsize
% \end{refsection}
\clearpage

% \newpage\textcolor{white}{.}\newpage\newpage\textcolor{white}{.}\newpage

\appendix
\renewcommand\thefigure{\thesection.\arabic{figure}}  
\renewcommand{\theHfigure}{A\arabic{figure}}
\setcounter{figure}{0}

\begin{appendices}

% \begin{refsection}
% Get the total number of entries in the first bibliography
% For the appendix accompanying our paper and videos demonstrating the method please visit our website:  \url{www.robot-learning.uk/adapting-skills}. \newpage

For videos demonstrating our method, please visit our website:  \href{https://www.robot-learning.uk/adapting-skills}{www.robot-learning.uk/adapting-skills}. For a pseudo-code explaining our method as presented in Section~\ref{sec:method} of the paper, please see  Algorithm~\ref{algo:train} for details on the self-supervised data collection and training process, Algorithm~\ref{algo:skill-learn} for details on skill acquisition and Algorithm~\ref{algo:skill-dep} for details on deploying and adapting skills with our method.
\section{Method}\label{sec:app-a}
In this Section, we provide further details regarding the training and implementation process of our network, $f_\theta$, as introduced in Section~\ref{sec:ref-grasp} of the paper. 

\begin{figure*}[h!]
    \centering
    \vspace{-1.39cm}\includegraphics[width=\textwidth]{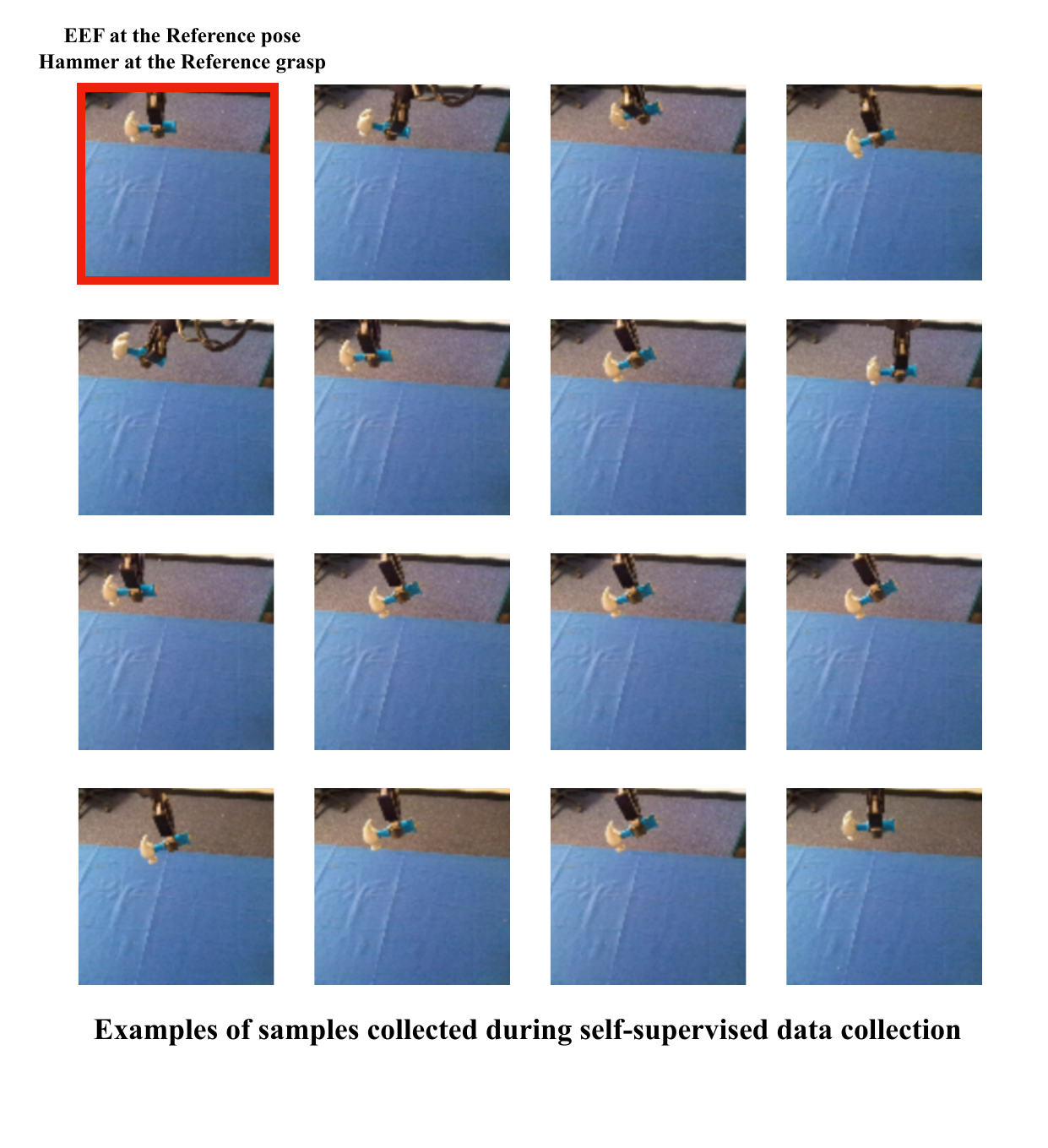}\vspace{-1.2cm}
    \caption{Images sampled from the dataset $\mathcal{D}$ collected in a self-supervised manner. The image on the top left (marked as red) shows the EEF at the reference pose with the object grasped at the reference grasp. While the EEF moves around the reference pose to emulate different arbitrary grasps, the object remains rigidly grasped at the reference grasp.}\vspace{.12cm}
    \label{fig:data-col}
\end{figure*}

Figure~\ref{fig:data-col} shows examples of images collected in the dataset $\mathcal{D}$ for the Hammer object. The EEF moves in a self-supervised manner to random poses around a reference pose $^\textrm{R}\mathbf{T}_{WE}$ to emulate different grasps with the object rigidly grasped at a reference grasp with pose $^\textrm{R}\mathbf{T}_{EO}$. Each image depicts how the grasped object may appear relative to the EEF if it is grasped this way when the EEF is at the reference pose $^\textrm{R}\mathbf{T}_{WE}$. 

For example, the EEF with the object grasped at the reference grasp in Figure~\ref{fig:emulates} (a) emulates the grasp shown in Figure~\ref{fig:emulates} (b) when the EEF is at the reference pose. Figure~\ref{fig:emulates} (c) emulates the grasp shown in Figure~\ref{fig:emulates} (d). The images in Figures~\ref{fig:emulates} (a) and (c) are part of our training dataset $\mathcal{D}$. As discussed in Section~\ref{sec:ref-grasp}, each image in the dataset has a transformation pair $^N\mathbf{T}^{-1}_{EE^{'}}$ that aligns the depicted grasp to the reference grasp, as discussed in Section~\ref{sec:ref-grasp}. 

During skill deployment, assume that the EEF at $^\textrm{R}\mathbf{T}_{WE}$ grasps the object as depicted in either Figure~\ref{fig:emulates} (b) or (d). For $f_\theta$ to correctly predict the corresponding $^N\mathbf{T}^{-1}_{EE^{'}}$ for each image, $f_\theta$ must be robust to the appearance of the background and EEF relative to the grasped object. This is because in the emulated grasps of Figures~\ref{fig:emulates} (a) and (c), the object always appears at a pose $^\textrm{R}\mathbf{T}_{EO}$ relative to the EEF, while in Figures~\ref{fig:emulates} (b) and (d) the object's appearance relative to the EEF is different. For this reason, we seek to segment the EEF and the background. Any segmentation method can be used to achieve this. In our work, we leverage the pretrained flow network of \cite{xu2022unifying} and deploy it in a similar manner to \cite{Boerdijk2020SelfSupervisedOS} to (1) train a segmentation network that segments the robot's EEF and (2) train an object-specific segmentation network, as follows.

\subsection{Dataset Segmentation}\label{sec:dataset-segmentation}
\subsubsection{EEF segmentation network}
\begin{figure*}[h!]
    \centering
\includegraphics[width=.8\textwidth]{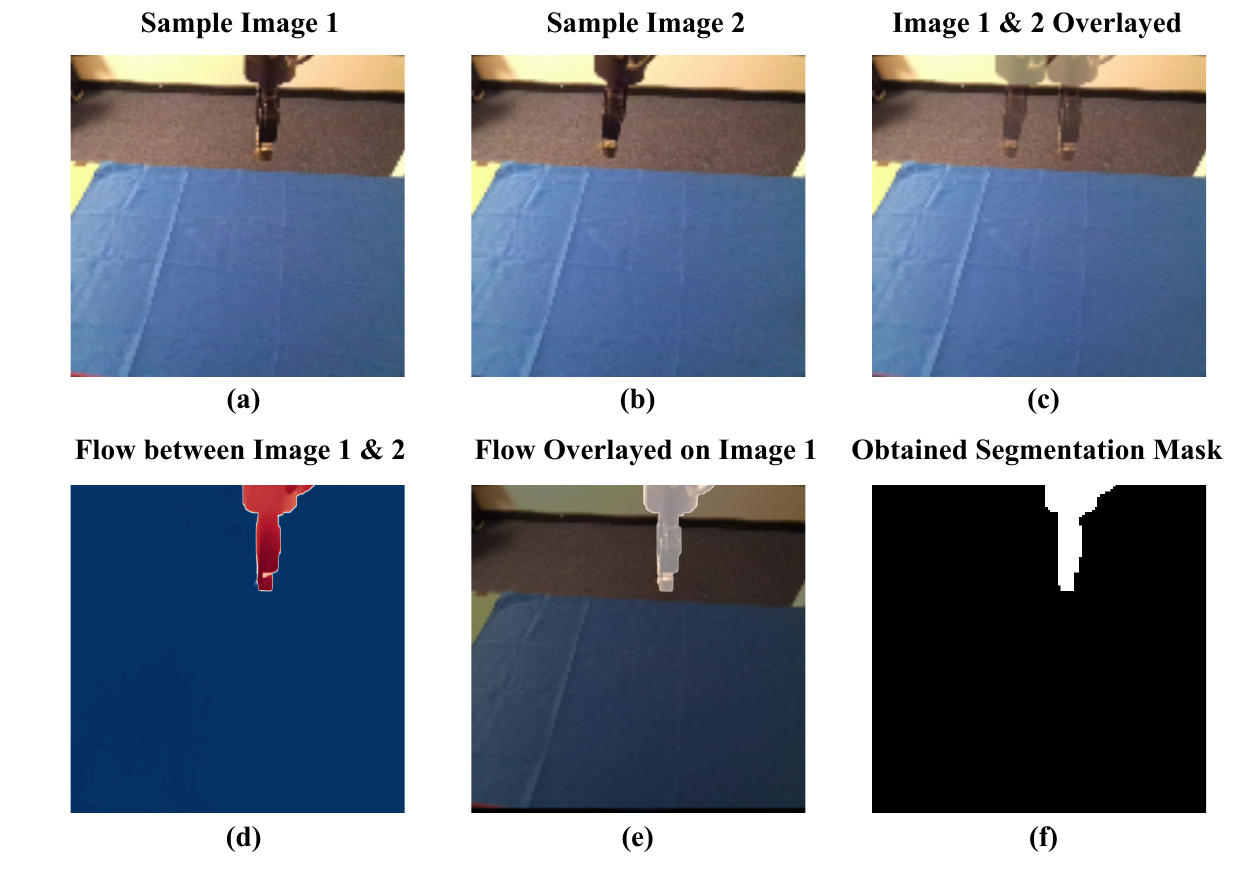}
    \caption{Figures (a) and (b) show two images sampled from the dataset collected after moving the EEF without any grasped object in front of the camera. Figure (c) shows Figure (a) and (b) overlayed to demonstrate the difference in the pose of the EEF between the two figures. Figure (d) shows the flow obtained between Figure (a) and (b) after being passed through the pretrained flow network of \cite{xu2022unifying}. Figure (e) shows the flow overlayed on top of the EEF of Figure (a). This allows us to obtain a segmentation mask for the EEF for Figure (a) as shown in Figure (f). This process is repeated over all the images collected for the EEF and allows us to train the EEF segmentation network.}        \label{fig:eef-flow}
\vspace{-.5cm}
\end{figure*}

First, we collect a dataset of images depicting the EEF at different poses by moving the EEF in a self-supervised manner to random poses in front of the external camera, with \textit{no} grasped object. Then, we use the flow network of \cite{xu2022unifying} to compute flow between pairs of images in the collected dataset. As the background is static, the computed flow allows us to obtain segmentation masks for the EEF in each image. An example of EEF images collected in the dataset, as well as the flow computed between them and the corresponding segmentation masks can be seen in Figure~\ref{fig:eef-flow}. Finally, given the collected dataset and segmentation masks, we train an EEF segmentation network. This process needs to be completed only \textit{once}, and the EEF segmentation network can be reused across any experiment and grasped object without further training.
\begin{figure*}[h!]
    \centering
    \includegraphics[width=\textwidth]{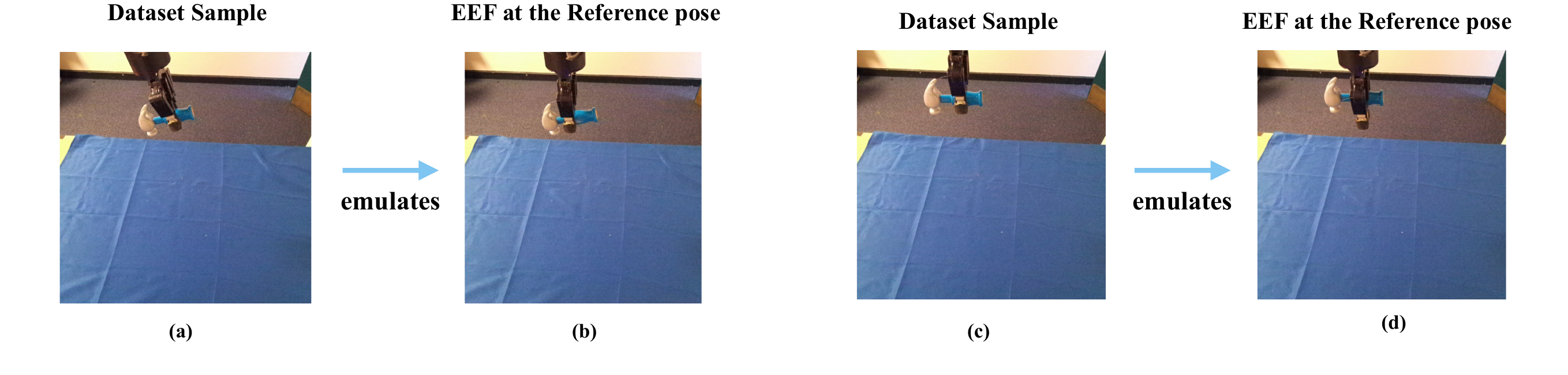}\vspace{-.49cm}
    \caption{Figures (a) and (c) are images sampled from dataset $\mathcal{D}$ collected during self-supervised data collection. Each image in (a) and (c) emulates a grasp depicted in the images of Figures (b) and (d) respectively where the EEF is at the reference pose. We manually reproduced the grasps in Figures (b) and (d) for visualisation.}\label{fig:emulates}\vspace{-.5cm}
    
\end{figure*}

\subsubsection{Object-specific segmentation network}
\begin{figure*}[t!]
    \centering
    \vspace{-.39cm}\includegraphics[width=.9\textwidth]{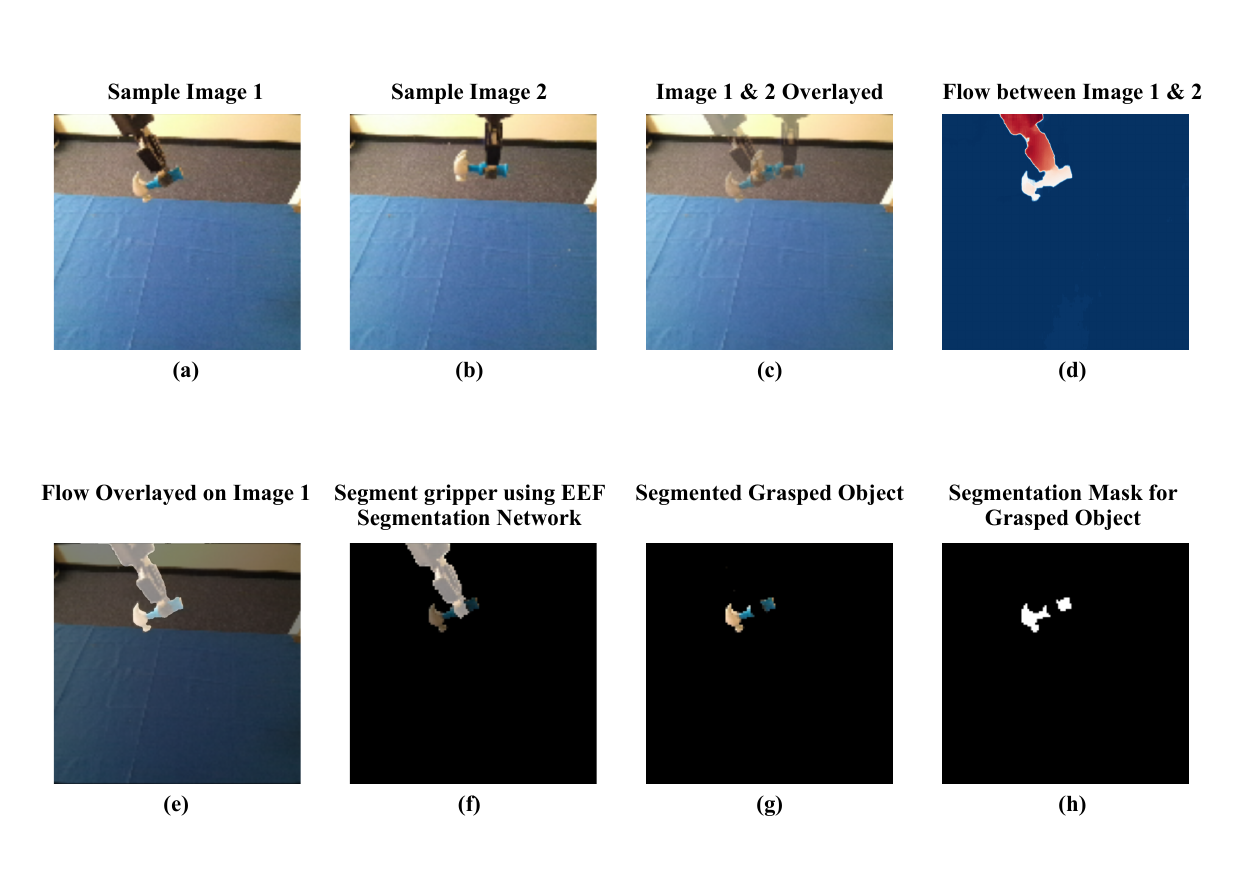}
    \vspace{-.89cm}\caption{Figures (a) and (b) demonstrate two images sampled from the dataset $\mathcal{D}$. Figure (c) overlays Figures (a) and (b) to demonstrate their difference. Figure (d) shows the flow obtained by passing Figures (a) and (b) through the pretrained flow network of \cite{xu2022unifying}. The flow network detects only the EEF and grasped object as the background is static. Figure (e) demonstrates the obtained flow overlayed on Figure (a). As there is no flow on the background, we remove it in Figure (f). Further, in Figure (f) we deploy the EEF segmentation network to detect and segment the EEF, leaving us only with the grasped object as shown in Figure (g). This allows us to obtain a segmentation mask only for the grasped object, as shown in Figure (h). This process is repeated over all the images collected in $\mathcal{D}$ and allows us to train an object-specific segmentation network.}\label{fig:object-flow}
    
\end{figure*}

To obtain a segmentation network for each grasped object, no further data collection is required, apart from the already collected dataset $\mathcal{D}$. Segmentation masks for the grasped object are obtained in an identical manner to the masks for the EEF segmentation network. For pairs of images in $\mathcal{D}$, the flow network detects both the EEF and the grasped object. Hence, we leverage our EEF segmentation network to remove the EEF, leaving us only with segmentation masks for the grasped object. Then, we train an object-specific segmentation network that receives an image in $\mathcal{D}$ and regresses the segmentation mask of the grasped object. Examples of this process can be seen in Figure~\ref{fig:object-flow}. This process needs to be repeated for every new object we need to adapt skills to, but it leverages the already collected dataset $\mathcal{D}$, requiring no additional time for data collection.

All our segmentation networks use the UNet \cite{uNet} network architecture.

\subsection{Dataset Augmentation}\label{sec:data-augmentation}
During data collection, a small part of the grasped object depicted in each image in $\mathcal{D}$ is not visible as the EEF's fingers occlude it. Additionally, as discussed in Section~\ref{sec:method}, each image depicts some arbitrary grasp under which the EEF may grasp the object when it is at the reference pose $^\textrm{R}\mathbf{T}_{WE}$. Hence, at test time (see Section~\ref{sec:ref-grasp}) if the EEF at $^\textrm{R}\mathbf{T}_{WE}$ grasps an object as depicted in one of the images in the dataset, a part of that object occluded in $\mathcal{D}$ will become visible, and a part of that object that is visible in the image in $\mathcal{D}$ will now be occluded by the EEF's fingers. This can be seen clearly in Figures~\ref{fig:emulates} (a) and (b). The part of the Hammer's handle occluded by the EEF's fingers in the emulated grasp of Figure~\ref{fig:emulates} (a) becomes visible when the Hammer is grasped in that pose when the EEF is at $^\textrm{R}\mathbf{T}_{WE}$ (Figure~\ref{fig:emulates} (b)). Further, the part of the Hammer's handle to the left of the EEF's fingers in the emulated grasp of Figure~\ref{fig:emulates} (a) is visible but becomes occluded in the grasp of Figure~\ref{fig:emulates} (b) by the EEF's fingers.

\begin{figure*}[h!]
    \centering
    \vspace{-.0cm}\includegraphics[width=\textwidth]{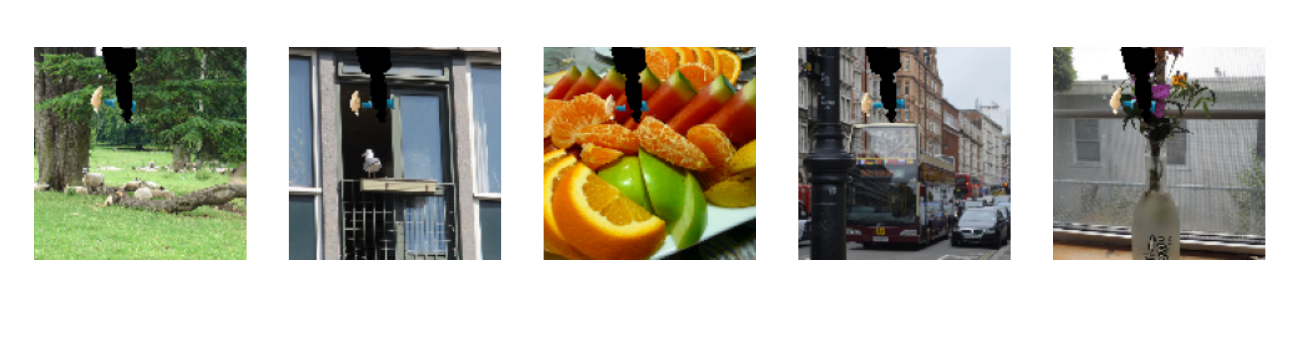}\vspace{-3ex}
    \caption{This Figure shows 5 images of the grasped object sampled from $\mathcal{D}$. Everything but the grasped object is segmented using the object-specific segmentation network we have trained. Then, we augment the background with MS-COCO \cite{lin2014microsoft} images for $f_\theta$ to learn to ignore the part of the grasped object that is occluded in $\mathcal{D}$ but becomes visible at test time. Further, we apply the segmentation mask of the EEF as it appears at the reference pose to occlude (segment) the part of the object that is visible in $\mathcal{D}$ but becomes occluded at test time when the EEF is at the reference pose due to the EEF's fingers.}\vspace{-.5cm}\label{fig:coco}
    
\end{figure*}

To make $f_\theta$ robust to this difference between each image collected in $\mathcal{D}$ and each image observed at test time, we need to ensure that $f_\theta$ (1) ignores the part of the object that was occluded in $\mathcal{D}$ but becomes visible at test time and (2) is robust to the fact that a part of the object visible in $\mathcal{D}$ becomes occluded at test time. To achieve (1), after segmenting the EEF and background from each image, we augment the background with MS-COCO \cite{lin2014microsoft} images. This way, $f_\theta$ learns to ignore the part of the object that was occluded in $\mathcal{D}$ but becomes visible at test time. To achieve (2), before data collection begins, we use our EEF segmentation network to obtain and store a segmentation mask of the EEF at the reference pose $^\textrm{R}\mathbf{T}_{WE}$. Then, we apply that segmentation mask to the images in $\mathcal{D}$ to occlude (segment) the part of the object that will be occluded at test time if the EEF at the reference pose $^\textrm{R}\mathbf{T}_{WE}$ grasps the object as depicted in the corresponding image. Figure~\ref{fig:coco} shows examples of this data augmentation process. Finally, to make $f_\theta$ robust to varying lighting conditions, we apply standard domain randomisation techniques such as varying the brightness or saturation of each image.
\subsection{Training}\label{sec:app-training}
To train $f_\theta$ for each object, first, we use the object-specific segmentation network to segment everything but the grasped object in each image in $\mathcal{D}$. Then, we train $f_\theta$ to receive as input each image in $\mathcal{D}$ and regress the corresponding $^N\mathbf{T}^{-1}_{EE^{'}}$ while applying the data augmentation and randomisation strategy of Section I.\ref{sec:data-augmentation}. As our objective function, we use the mean squared error loss and the Adam \cite{kingma2014method} optimiser. In practice, we train two versions of $f_\theta$, one that regresses all DoFs relating to position and one that regresses all DoFs relating to orientation. We found this to perform better than training a single network to regress all of $SE(3)$ directly. As discussed in Section~\ref{sec:experiments}, $f_\theta$ is trained over random poses sampled around $^RT_{WE}$ in the range of 30cm for each of the DoFs relating to position and 60$^\circ$ for each of the DoFs relating to orientation. However, our method is not limited to a particular range around the reference pose and can be trained over any desirable range of poses. We found that 30cm and 60$^\circ$ were enough to cover the majority of skill and deployment grasps possible for the objects we used in our experiments (Section~\ref{sec:experiments}, Figure~\ref{fig:everyday-objects}). Finally, we also train a function with parameters $\psi$, $g_\psi:\mathbb{R}^{H\times W\times C}\rightarrow SE(3)$ ($H$ corresponds to image height, $W$ to image width and $C$ to image channel) in an identical manner to $f_\theta$ but on poses sampled over a shorter range around the reference pose (6cm for each of the DoFs relating to position and 12$^\circ$ for each of the DoFs relating to orientation). No extra data collection is necessary to train $g_\psi$, it is trained simply on a smaller subset of $\mathcal{D}$. We found that doing so improves the accuracy of our method. In practice, we deploy $f_\theta$ and $g_\psi$ sequentially in that order.

\subsection{Network Architecture}\label{sec:net-arch}
For our networks' architectures, we use a UNet \cite{uNet} encoder with CoordConv \cite{cOOrdconv} layers, and self-attention \cite{sagan} followed by a simple MLP that outputs a per DoF prediction. We found this network architecture to perform better when compared to replacing the UNet encoder with a pre-trained ResNet \cite{he2016residual}, a Deep Spatial Autoencoder \cite{dsae} and a simple CNN. We also found that for the UNet encoder, using CoordConv layers and self-attention compared to standard convolutions improved our network's performance, but not significantly.

\subsection{Data Collection Time}\label{sec:data-col-time}

As noted in our experiments section, we typically allocate around 5 minutes for data collection, during which the robot moves in front of the external camera. In practice, varying data collection times have different effects on our method's performance. Overall, we noted that increasing data collection time resulted only in a slight performance increase. The most important criterion for high accuracy was collecting enough views to cover the grasped object from all sides. Consequently, shortening the data collection time would result in lower accuracy as fewer object views would be collected. On the other hand, however, if we were operating a robot that can move faster and more accurately, then we could collect the same amount of data as we did in $5$ minutes, but in a shorter amount of time. Hence, determining the right data collection time depends on our robotic hardware, but as long as the collected dataset contains views covering the whole object then performance is expected to be high.

 \subsection{Visual servoing to align grasps to the reference grasp}\label{sec:visual-servoing}
As discussed in Section~\ref{sec:alignment-network}, in practice, we deploy our method in a visual servoing (VS) manner. However, as the predictions made by $f_\theta$ (and $g_\psi$) are made relative to the reference pose $^\textrm{R}\mathbf{T}_{WE}$, they are not directly amenable to VS. For this reason, we need to transform them appropriately at each step of the VS process

% \footnote{For all the equations of Section~\ref{sec:visual-servoing} when aligning a skill grasp (instead of a deployment grasp) to the reference grasp, simply replace  ${^{\textrm{D}}}\mathbf{T}_{EE^{'}}$ with ${^{\textrm{S}}}\mathbf{T}_{EE^{'}}$ as they are equivalent (see Section~\ref{sec:method} of the paper).}.

The prediction of $f_\theta$ provides us with a single transformation to align any deployment or skill grasp to the reference grasp. Instead of making a single prediction, it can be beneficial to start moving the EEF to the predicted pose $^N\mathbf{T}^{-1}_{EE^{'}}$ relative to $^R\mathbf{T}_{WE}$ and, at the same time, leverage new live images captured from the camera to make further predictions with our network in a VS manner. However, the predictions made by $f_\theta$ are valid only when the EEF is in front of the camera at the reference pose $^\textrm{R}\mathbf{T}_{WE}$. Hence, as the EEF begins to move from the reference pose $^\textrm{R}\mathbf{T}_{WE}$ to the predicted pose $^R\mathbf{T}_{WE}{^N}\mathbf{T}^{-1}_{EE^{'}}$, we cannot directly apply new predictions made by $f_\theta$ based on live images from the external camera. However, we can still leverage these predictions by accounting for the EEF's pose at the timestep those predictions were made, as follows. 
First, we denote as ${}^{t=\lambda}{^N}\mathbf{T}^{-1}_{EE^{'}}
$ to be the prediction made by $f_\theta$ based on the image $I_{t=\lambda}$ captured at timestep $t=\lambda$ from the camera. Then, to incorporate ${}^{t=\lambda}{^N}\mathbf{T}^{-1}_{EE^{'}}$ under a VS framework, at timestep $\lambda$ we can apply a transformation ${{{^{VS}}}\mathbf{T}}_{EE^{'}}^{t=\lambda}$ in the EEF's frame $\{E\}$, where:\begin{equation}\label{eq9}
    {{{^{VS}}}\mathbf{T}}_{EE^{'}}^{t=\lambda}= \big[{{^{t=\lambda}}}\mathbf{T}_{EE^{'}}^{-1}\big]\big[{}^{t=\lambda}{^N}\mathbf{T}^{-1}_{EE^{'}}
\big]\big[{{^{t=\lambda}}}\mathbf{T}_{EE^{'}}\big]\,\,,
\end{equation}where ${{^{t=\lambda}}}\mathbf{T}_{EE^{'}}$ corresponds to the pose of the EEF relative to $^\textrm{R}\mathbf{T}_{WE}$ at timestep $\lambda$, that is $${{^{t=\lambda}}}\mathbf{T}_{EE^{'}} = {^\textrm{R}}\mathbf{T}_{WE}^{-1}{^{t=\lambda}}\mathbf{T}_{WE}.$$ Further, at time-step $t=0$, $ {\mathbf{T}}_{VS}^{t=0}={}^{t=0}{^N}\mathbf{T}^{-1}_{EE^{'}}
$, as the EEF is at $^R\mathbf{T}_{WE}$ and $^{t=0}\mathbf{T}_{EE^{'}}=\mathbf{I}$, where $\mathbf{I}$ is the identity matrix. If we stopped the VS process at timestep $t=0$, this would be identical to making a single prediction with $f_\theta$. By performing VS, we can iteratively leverage predictions made by $f_\theta$ as more live images are captured from our camera either for a fixed amount of time or until our network predicts the identity matrix, in which case the EEF has been transformed such that the grasped object is aligned to the reference grasp. At the end of the VS process, ${^N}\mathbf{T}^{-1}_{EE^{'}}
$ simply equals the relative pose between the EEF at the reference pose and the EEF at the final timestep of the VS process. That is, assume we run VS for $\Lambda$ timesteps. Then, $${^N}\mathbf{T}^{-1}_{EE^{'}}
=\big[{^\textrm{R}}\mathbf{T}_{WE}^{-1}\big]\big[{^{t=\Lambda}}\mathbf{T}_{WE}\big].$$ As discussed in our Experiments (Section~\ref{sec:exp-accuracy}), when deploying our method, we perform VS for 5 seconds, where we allocate the first 2.5 seconds to the network $f_\theta$ and the last 2.5 seconds to the network $g_\psi$. For an algorithm showing the application of VS to align grasps to the reference grasp see Algorithm~\ref{algo:skill-dep}.
\newpage

\input{algorithm1.tex}

\input{algorithm2.tex}
\input{algorithm3.tex}

\section{Experiments}
\setcounter{table}{0}
\setcounter{figure}{0} 
\renewcommand{\theHfigure}{B\arabic{figure}}
\renewcommand{\theHtable}{B\arabic{table}}

\renewcommand{\thetable}{B\arabic{table}}

\subsection{Accuracy Evaluation using forward kinematics}\label{sec:acc-eval-proc}

As we have no access to the objects' 3D CAD models we cannot directly estimate their pose to evaluate our method's and the baselines' accuracy in obtaining the corrective transformation. Hence, as discussed in section~\ref{sec:exp-accuracy} we use the robot's forward kinematics as follows.

First, we move the robot to the reference pose and hand the object to the EEF at a random pose. This defines a potential skill grasp. We then deploy our alignment network to compute ${^{\textrm{S}}\mathbf{T}_{EE'}}$ (recall from Section~\ref{sec:corr-trans} that ${^{\textrm{S}}\mathbf{T}_{EE'}}$ is the transformation that aligns the skill grasp to the reference grasp and is identical to the predicted ${^{\textrm{N}}\mathbf{T}^{-1}_{EE'}}$ by our alignment network, only denoted ${^{\textrm{S}}\mathbf{T}_{EE'}}$ for clarity). \textit{Without} changing the grasp, we move the EEF to a random pose to emulate a possible deployment grasp. Then, similarly to the skill grasp, we use the alignment network to obtain the transformation that aligns the deployment grasp to the reference grasp, denoted ${^{\textrm{D}}\mathbf{T}_{EE'}}$. Using ${^{\textrm{S}}\mathbf{T}_{EE'}}$ and ${^{\textrm{D}}\mathbf{T}_{EE'}}$ we compute the corrective transformation as discussed in Section~\ref{sec:corr-trans}. If the corrective transformation is accurate, the EEF should return to the reference pose to align the deployment grasp to the skill grasp. By computing the error to the reference pose using the robot's forward kinematics we can  quantify the error in the obtained corrective transformation. During evaluation, the forward kinematics are used only for evaluation and are \textit{not} accessible to our method or any of the baselines.

\subsection{Baselines}\label{sec:baselines}

\textit{\textbf{ICP}}: Given a skill grasp, we store a segmented point cloud of the grasped object captured from the camera. To obtain the segmented point cloud, we deploy the same segmentation method used for our method. Then, given any deployment grasp, for each step of the visual servoing process, we capture a segmented point cloud and use ICP to compute the corrective transformation. We initialise ICP with the identity transformation and use the open-source implementation provided by Open3D \cite{open3d}, which we optimise for our tasks. %We also experimented with a variant of the ICP baseline, where instead of capturing segmented point clouds directly visible to the camera, at every timestep of the visual servoing process, we rotated the EEF by $180^{\circ}$ to capture both sides of the grasped object. The goal of this process was to get a more complete 3D CAD model of the object. However, we noticed that doing so was less accurate.

\textit{\textbf{Colour-ICP}}: We deploy Colour-ICP (C-ICP) in an identical manner to ICP and used the implementation of \cite{park2017colored} provided by Open3D \cite{open3d}.
%We also evaluated whether rotating the EEF to get a more complete 3D CAD model of the object improved    performance but noted that it was less accurate. 
%For Colour-ICP, we used the open-source implementation provided by Open3D  \cite{open3d}, which follows the one proposed in \cite{park2017colored}.

\textit{\textbf{real-world NDF (RW-NDF)}}: We train a variant of the Neural Descriptor Fields (NDFs) \cite{simeonovdu2021ndf}, on a segmented point cloud of the grasped object. This way, we test whether leveraging learned point cloud descriptors results in better performance when finding correspondences compared to the data association method used by ICP and C-ICP for our problem. To train NDFs in the real-world, we use the pre-trained network provided by the authors, which we fine-tune on the real-world point clouds for our grasped object. We found that fine-tuning the pre-trained network performed better than training on the observed point clouds from scratch. This way, we avoid the need to pre-train in simulation, which assumes prior knowledge of the object's category. Specifically, we fine-tuned a pre-trained occupancy network to reconstruct the volume near the point cloud of the grasped object by aggregating point clouds captured from two sides of the object by rotating the gripper by $180^\circ$ in front of the camera. During deployment, we followed the optimisation strategy described in~\cite{simeonovdu2021ndf}.

\textcolor{black}{\textbf{DINO:} Given a skill grasp we store a RGB and depth image. Then, both images are segmented using the same segmentation method used for our method. For any deployment grasp, we also capture and segment a pair of RGB and depth images. We then leverage the pretrained DINO ViT provided by the authors \cite{Caron2021EmergingPI} to establish correspondences between the RGB images of the skill and the deployment grasp as proposed in \cite{amir2021deep}. The fact that the RGB images depict only the grasped object after segmentation allows us to ensure that correspondences are established only between the grasped object in both the skill and deployment grasp images. Given the established correspondences, we then leverage the depth images to determine the corrective transformation using singular value decomposition (SVD) \cite{Arun1987LeastSquaresFO}. Sometimes DINO required several seconds to determine correspondences, in which case we did not limit DINO's deployment time to 5 seconds as we did for our methods. }

\textcolor{black}{\textbf{AspanFormer:} We deploy AspanFormer\cite{chen2022aspanformer} in an identical manner to the DINO baseline, but in order to establish correspondences between the skill and deployment grasp RGB images we use the pretrained model provided by the authors \cite{chen2022aspanformer}.}

% For the ICP and C-ICP baselines, we also experimented with rotating the EEF in front of the camera (similarly to what we did for RW-NDF) to obtain an aggregated point cloud of the grasped object. The goal behind this experiment was to test ICP and C-ICP after obtaining a more complete 3D model of the grasped object. However, we noticed that this resulted in lower performance and hence decided to deploy ICP and C-ICP as explained above. 

\subsection{Accuracy Results}
\begin{figure*}[h!]
    \captionsetup{labelfont={color=black}}
    \vspace{-.3cm}\centering
    \includegraphics[width=\textwidth]{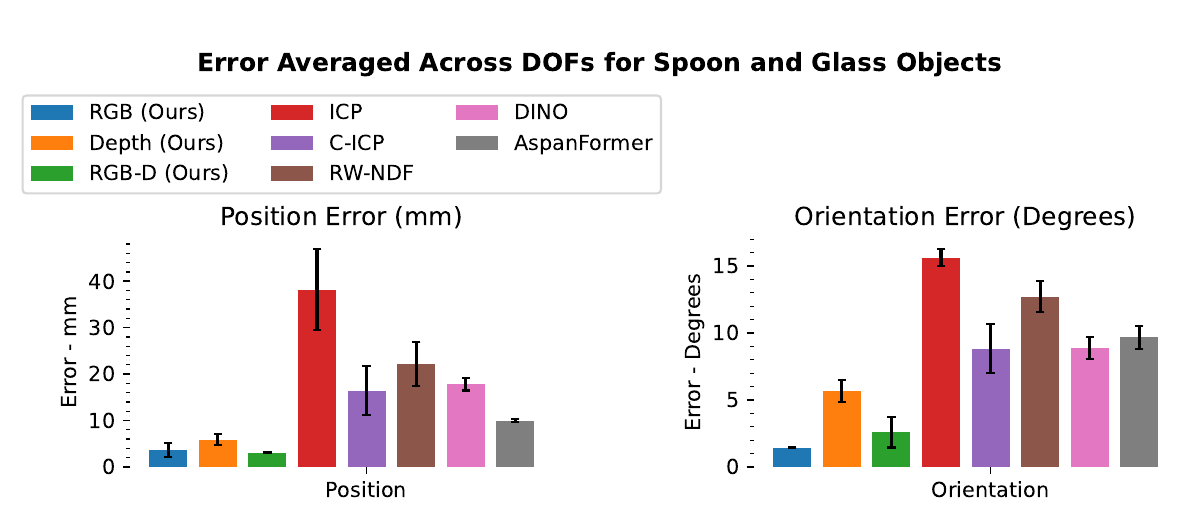}\vspace{-.27cm}
    
    \caption{\textcolor{black}{Mean and standard deviation error in computing the corrective transformation between grasps for the Spoon and Glass objects averaged across all DoFs for the position and orientation.}}
    \label{fig:accuracy-spoon-glass}
\end{figure*}

\begin{figure*}[h!]
    \captionsetup{labelfont={color=black}}
    \centering
    \includegraphics[width=\textwidth]{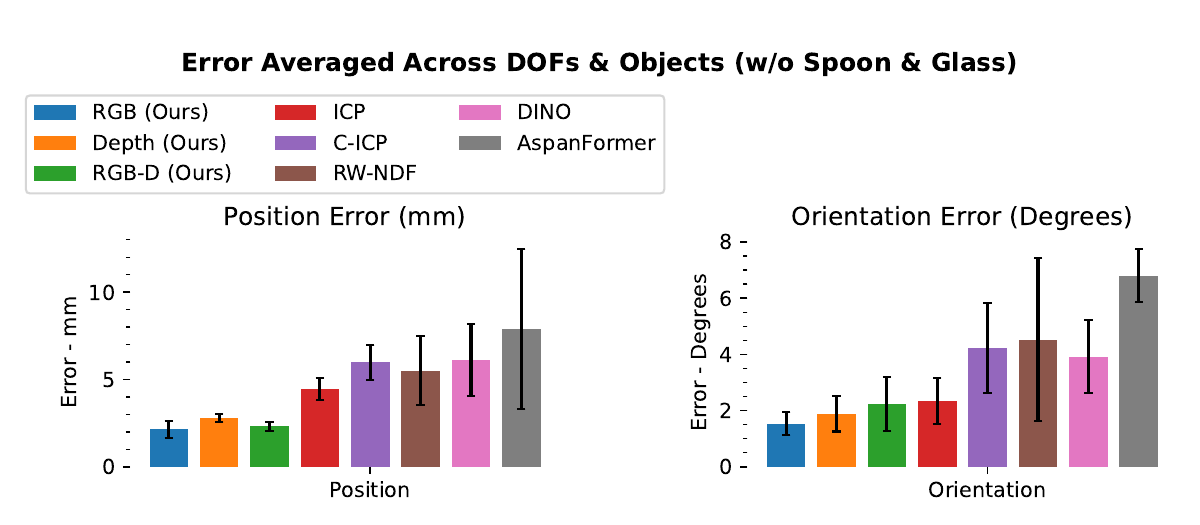}\vspace{-.27cm}
    \caption{\textcolor{black}{Mean and standard deviation error in computing the corrective transformation between grasps for the Hammer, Screwdriver, Wrench and Bread objects averaged across all DoFs for the position and orientation .}}
    \label{fig:accuracy-no-spoon-glass}\vspace{-.4cm}
\end{figure*}

The numerical values of Figure~\ref{fig:accuracy-main-results} of section I.\ref{sec:acc-eval-proc} corresponding to the mean and standard deviation error for the corrective transformation can be seen in Table~\ref{table:reorient-accuracy}. The corrective transformation mean and standard deviation error averaged across the Spoon and Glass objects can be seen in Figure~\ref{fig:accuracy-spoon-glass}. The baselines obtain a particularly low performance for these objects as their depth quality is low due to their shiny (spoon) and semi-transparent (glass) appearance. Figure~\ref{fig:accuracy-no-spoon-glass} shows the mean and standard deviation error on the corrective transformation for the rest of the objects (excluding the Spoon and Glass). As shown, for these objects the baselines perform significantly better compared to the Spoon and Glass objects but on average still worse when compared to all variants of our method.

\input{accuracy_tab}


\begin{figure}[t!]
\begin{minipage}{.48\textwidth}
     \centering
     \includegraphics[width=\linewidth]{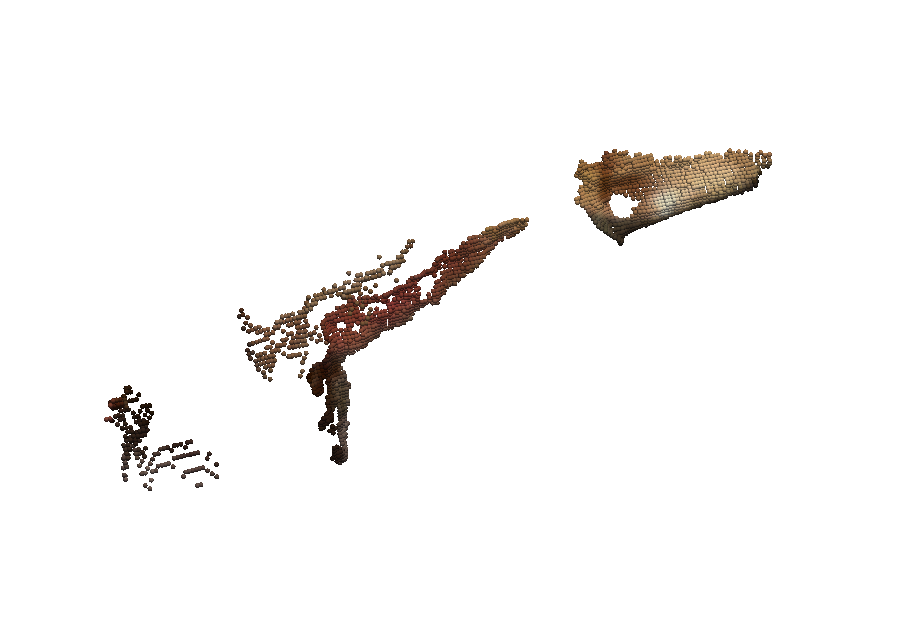}
     \caption{Point cloud of the spoon object. As shown the point cloud's quality is poor due to missing depth data.}\vspace{-2.5ex}\label{fig:spoon-pcd}
   \end{minipage}\hfill
   \begin{minipage}{0.48\textwidth}
     \centering
     \includegraphics[width=\linewidth]{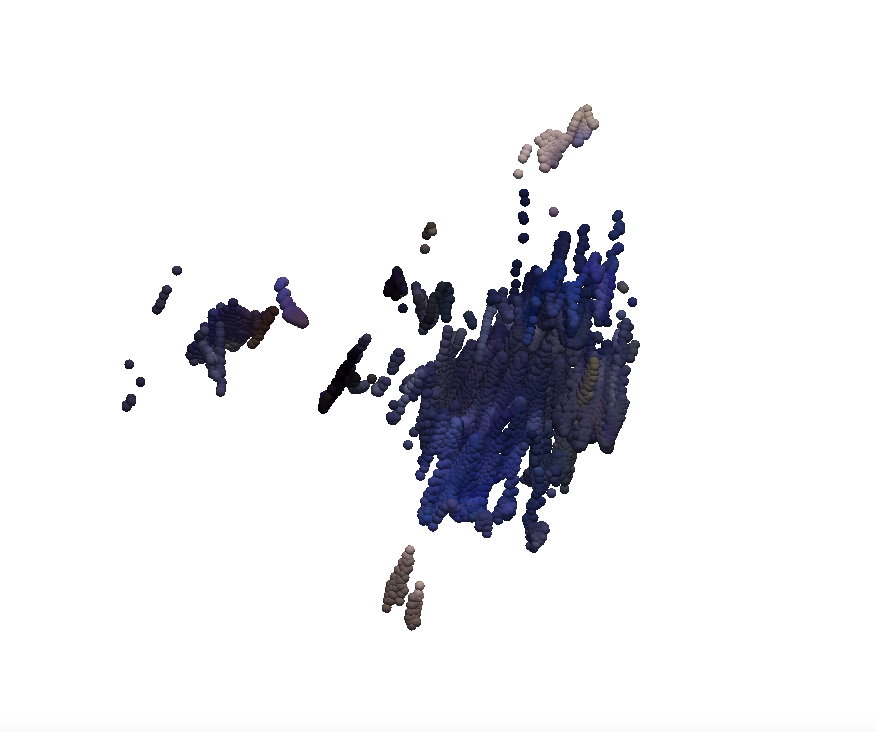}
     \caption{Point cloud of the glass object. As shown the point cloud's quality is poor due to missing depth data.}\vspace{-2.5ex}\label{Fig:glass-pcd}
   \end{minipage}
\end{figure}

\subsection{DOME}\label{sec:dome}
DOME \cite{valassakis2022dome} is a one-shot imitation learning method that enables efficient acquisition of robotic skills from a single demonstration. DOME comprises two parts: 1) a pre-trained visual servoing network that allows it to approach any target object specified in the demonstration (e.g., the nail for the Hammer task in Figure~\ref{fig:everyday-objects}) on a table-top setup regardless of its pose using a wrist camera rigidly mounted on the robot's EEF. The pre-trained visual servoing network can be deployed immediately to any setup for which we have provided a demonstration. 2) an interaction trajectory that is demonstrated to the robot by a human and defines how the robot interacts with objects in the workspace. That interaction trajectory consists of a sequence of twists tracked by the EEF which are recorded during the human demonstration. To generalize a demonstration across different configurations of the target object DOME uses an eye-in-hand camera. During skill deployment, DOME first approaches the target object using the observations made by the eye-in-hand and replays the demonstrated sequence of velocities (and as a result EEF poses). For further details we refer the reader to~\cite{valassakis2022dome}.

In DOME, for skills involving the manipulation of grasped objects, each skill is tailored to the specific pose the grasped object had in the EEF during the demonstration. Hence, if the object is grasped at a different pose, the demonstrated skill is no longer suitable to complete its designated task. Hence, we deployed our method to adapt DOME's skills to novel deployment grasps as discussed in the Experiments section~\ref{sec:dome-evaluation}.

% \printbibliography[title={References Appendix}]
% \end{refsection}

 \end{appendices}

\end{document}

%% file: peg_hole_table.tex
\begin{table}[]

\small
\setlength{\tabcolsep}{1pt}
\centering

% \addtolength{\leftskip} {-0.1cm} 

\vspace{1.5ex}\begin{tabular}{l|>{\centering}p{1cm}|>{\centering}p{1cm}|>{\centering}p{1.5cm}|>{\centering}p{1cm}|>{\centering\arraybackslash}p{1.5cm}}
\hline\textbf{Hole Tolerance   } & \textbf{RGB} & \textbf{Depth} &  \textbf{RGB-D}   &  \textbf{\textcolor{black}{ICP}}  &  \textbf{\textcolor{black}{DINO}}\\ \hline
\textbf{2mm} & \textbf{80\%} & 40\% & 40\%& \textcolor{black}{40\%}& \textcolor{black}{60\%} \\
\textbf{4mm} & \textbf{100\%} & 60\% & 80\% & \textcolor{black}{40\%}& \textcolor{black}{20\%}\\
\textbf{8mm} & \textbf{100\%} & 80\% & \textbf{100\%}& \textcolor{black}{\textbf{100\%}}& \textcolor{black}{80\%} \\
\textbf{12mm} & \textbf{100\%} & 100\% & \textbf{100\%}& \textcolor{black}{\textbf{100\%}}& \textcolor{black}{\textbf{100\%}} \\ \hline
\textbf{Average} & \textbf{95\%} & 70\% & {80}& \textcolor{black}{{70\%}}& {{65\%}} \\ \hline

\end{tabular}\caption{Skill adaptation success rate to various deployment grasps for peg-in-hole insertion for 4 hole tolerances. \label{table:peh-in-success-rate}\vspace{-3.5ex}}
\end{table}
% \vspace{-.47cm}

%% file: dome_results.tex
\begin{table}[]

\small
\setlength{\tabcolsep}{1pt}
\centering

% \addtolength{\leftskip} {-0.1cm} 

\vspace{1.5ex}\begin{tabular}{l|>{\centering}p{1cm}|>{\centering}p{1cm}|>{\centering}p{1.5cm}|>{\centering}p{1cm}|>{\centering\arraybackslash}p{1.5cm}}
\hline\textbf{Tasks   } & \textbf{RGB} & \textbf{Depth} &  \textbf{RGB-D}   &  \textcolor{black}{\textbf{ICP}}  &  \textcolor{black}{\textbf{DINO}}\\ \hline
\textbf{Hammer} & \textbf{100\%} & \textbf{100\%} & \textbf{100\%} & \textcolor{black}{80\%}& \textcolor{black}{60\%}\\
\textbf{Screwdriver} & \textbf{30\%} & 10\% & 10\% & \textcolor{black}{20\%} & \textcolor{black}{0\%} \\
\textbf{Bread} & 50\% & 60\% & \textbf{70\%} & \textcolor{black}{60\%} & \textcolor{black}{\textbf{70}\%}\\ 
\textbf{Spoon} & \textbf{100\%} & \textbf{100\%} & \textbf{100\%}& \textcolor{black}{30\%}& \textcolor{black}{40\%} \\
\textbf{Wrench} & 70\% & 70\% & \textbf{80\%}& \textcolor{black}{50\%}& \textcolor{black}{20\%} \\
\textbf{Glass} & \textbf{100\%} & 80\% & 80\% & \textcolor{black}{20\%}& \textcolor{black}{40\%}\\\hline
\textbf{Average} & \textbf{75\%} & 70\% & 73\% & \textcolor{black}{43\%}& \textcolor{black}{38\%}\\\hline

\end{tabular}\caption{Skill adaptation success rate to various deployment grasps for imitation learning skills taught using DOME \label{table:dome-results}}\vspace{-4.2ex}
\end{table}

%% file: algorithm1.tex
\vspace{-20cm}\begin{algorithm*}[h!]
\SetAlgoLined
\KwIn{Reference pose $^\textrm{R}\mathbf{T}_{WE}$, Reference grasp $^\textrm{R}\mathbf{T}_{EO}$, Training dataset $\mathcal{D}=\{\}$, Segmentation dataset $\mathcal{S}=\{\}$ \tcp{$^\textrm{R}\mathbf{T}_{WE}$ can be any pose as long as the EEF is visible to the camera. $^\textrm{R}\mathbf{T}_{EO}$ can be any grasp.}} 
\begin{algorithmic}[1]
\FOR{\(\mbox{iteration } m=1 \textrm{ to } M\)} 
\STATE Sample pose $^\textrm{N}\mathbf{T}_{EE^{'}}$ \tcp{see Section~\ref{sec:ref-grasp}}
\STATE Move EEF to $^\textrm{R}\mathbf{T}_{WE}{^\textrm{N}}\mathbf{T}_{EE^{'}}$
\STATE Capture image $I$ from external camera
\STATE Store sample: $\mathcal{D}=\mathcal{D}\cup \{(I, {^N}T^{-1}_{EE^{'}})_{m}\}$
\ENDFOR

\SetNoFillComment
\FOR{\(\mbox{iteration } m=1 \textrm{ to } M-1\)}
\STATE Sample pairs of images: $(I_m, I_{m+1})\sim\mathcal{D}$
\STATE Compute flow between the pair of images $(I_m, I_{m+1})$ using the pretrained flow network of \cite{xu2022unifying}
\STATE Segment EEF and background and obtain object mask $b_m$ \tcp{see Section~\ref{sec:dataset-segmentation}}
\STATE Store sample: $\mathcal{S}=\mathcal{S}\cup(I_m, b_m)$
\ENDFOR
\STATE Train object-specific segmentation network on $\mathcal{S}$
\FOR{\(\mbox{iteration } m=0 \textrm{ to } M\)}
\STATE Remove EEF and background from all images in $\mathcal{D}$ using the object-specific segmentation network
\ENDFOR
\STATE Train $f_\theta$ on data augmented images of $\mathcal{D}$ \tcp{see Section~\ref{sec:data-augmentation}}
\STATE Train $g_\psi$ on data augmented images of $\mathcal{D}$ \tcp{see Section~\ref{sec:data-augmentation}}
% \STATE Learn skill $\mathbf{S}$ starting from any desirable EEF pose, with any desirable object grasp using any method (see Section V.D of the paper for assumptions on skills)
\caption{Self-supervised Data Collection \& Training}\label{algo:train}
\end{algorithmic}{}
\KwOut{$f_\theta, g_\psi$}

\end{algorithm*}

%% file: algorithm2.tex
\begin{algorithm*}[h!]
\SetAlgoLined
\KwIn{Networks $f_\theta$, $g_\psi$, Object specific segmentation network}
\begin{algorithmic}[1]
\STATE \textbf{Skill grasp. }Grasp object at any desirable pose in the EEF suitable to learn the desirable skill
\STATE Move the EEF to $^\textrm{R}\mathbf{T}_{WE}$
\SetNoFillComment
\FOR{\(\mbox{iteration } \lambda=0 \textrm{ to } \Lambda\)}
\STATE Obtain live image $I$ from external camera
\STATE Remove EEF and background using the object specific segmentation network
\IF{$t \leq \Lambda/2$}
\STATE Obtain $^\textrm{S}\mathbf{T}_{EE^{'}}=f_\theta(I)$ \tcp{Aligns skill grasp to the reference grasp using the alignment network;  see Section~\ref{sec:corr-trans} and I.~\ref{sec:app-training}}
\ELSIF{$t > \Lambda/2$}
\STATE Obtain $^\textrm{S}\mathbf{T}_{EE^{'}}=g_\psi(I)$ \tcp{Aligns skill grasp to the reference grasp using the alignment network; see Section ~\ref{sec:corr-trans} and I.~\ref{sec:app-training}}
\ENDIF
\STATE Move the EEF to: $\big[{^{t=\lambda}}\mathbf{T}_{WE}\big]\big[{{{^{VS}}}\mathbf{T}}_{EE^{'}}^{t=\lambda}\big]$ \tcp{see Section I.\ref{sec:visual-servoing}}
\ENDFOR
\STATE Obtain $^\textrm{S}\mathbf{T}_{EE^{'}}=\big[{^R}\mathbf{T}_{WE}^{-1}\big]\big[{^{t=\Lambda}\mathbf{T}_{WE}}\big]$ \tcp{see Section I.\ref{sec:visual-servoing}}
\STATE Define skill trajectory $\mathbf{S}$ starting from any desirable initial EEF pose, with a desirable skill acquisition method. 

\caption{Skill Acquisition}\label{algo:skill-learn}
\end{algorithmic}{}
\KwOut{$^\textrm{S}\mathbf{T}_{EE^{'}}$, acquired skill trajectory $\mathbf{S}$}
\end{algorithm*}

%% file: algorithm3.tex
\begin{algorithm*}[h!]
\SetAlgoLined
\KwIn{$^\textrm{S}\mathbf{T}_{EE^{'}}$, acquired skill $\mathbf{S}$}
\begin{algorithmic}[1]
\STATE \textbf{Deployment grasp. }Grasp the object at any desirable pose in the EEF to deploy the learned skill $\mathbf{S}$
\STATE Move the EEF to $^\textrm{R}\mathbf{T}_{WE}$
\SetNoFillComment
\FOR{\(\mbox{iteration } \lambda=0 \textrm{ to } \Lambda\)}
\STATE Obtain live image $I$ from external camera
\STATE Remove EEF and background using the object-specific segmentation network
\IF{$t \leq \Lambda/2$}
\STATE Obtain $^\textrm{D}\mathbf{T}_{EE^{'}}=f_\theta(I)$  \tcp{Aligns deployment grasp to the reference grasp using the alignment network;  see Section~\ref{sec:corr-trans} and I.~\ref{sec:app-training}}
\ELSIF{$t > \Lambda/2$}
\STATE Obtain $^\textrm{D}\mathbf{T}_{EE^{'}}=g_\psi(I)$ \tcp{Aligns deployment grasp to the reference grasp using the alignment network;  see Section~\ref{sec:corr-trans} and I.~\ref{sec:app-training}}
\ENDIF
\STATE Move the EEF to: $\big[{^{t=\lambda}}\mathbf{T}_{WE}\big]\big[{{{^{VS}}}\mathbf{T}}_{EE^{'}}^{t=\lambda}\big]$ \tcp{see Section~ I.\ref{sec:visual-servoing}}
\ENDFOR
\STATE Obtain $^\textrm{D}\mathbf{T}_{EE^{'}}=\big[{^R}\mathbf{T}_{WE}^{-1}\big]\big[{^{t=\Lambda}\mathbf{T}_{WE}}\big]$ \tcp{see Section I.\ref{sec:visual-servoing})}
\STATE Compute corrective transformation $^\textrm{C}\mathbf{T}_{EE^{'}}={{^\textrm{S}}}\mathbf{T}_{EE^{'}}^{-1}{^\textrm{D}}\mathbf{T}_{EE^{'}}$ \tcp{see Section~\ref{sec:corr-trans}}
\STATE Deploy and adapt skill from any desirable EEF pose using the corrective transformation $^\textrm{C}\mathbf{T}_{EE^{'}}$ \tcp{see Section~\ref{sec:problem-formulation}}
\caption{Skill Deployment \& Adaptation}\label{algo:skill-dep}
\end{algorithmic}{}
\KwOut{Adapted skill executed}
\end{algorithm*}

%% file: accuracy_tab.tex
\begin{table*}[t!]
\renewcommand{\arraystretch}{1.2}

\tiny
\setlength{\tabcolsep}{1pt}
\centering

\caption{Mean and standard deviation of  the corrective transformation error using the robot's forward kinematics\label{table:reorient-accuracy}}
% \addtolength{\leftskip} {-0.1cm} 
\resizebox{\textwidth}{!}{
\begin{tabular}{ccccccccccccccccccccccccccc}
& \multicolumn{2}{c}{\textbf{RGB (ours)}}   & \multicolumn{2}{c}{\textbf{Depth (ours)}}   & \multicolumn{2}{c}{\textbf{RGB-D (ours)}} & \multicolumn{2}{c}{\textbf{ICP}} & \multicolumn{2}{c}{\textbf{Color ICP} }                                                                                                                                   & \multicolumn{2}{c}{\textbf{RW-NDF}  }                                                                                                             
\\ \cline{1-13} 
                                   \multicolumn{1}{c|}{\textbf{Objects}} & \begin{tabular}[c]{@{}c@{}}mm\end{tabular} & \multicolumn{1}{c|}{\begin{tabular}[c]{@{}c@{}}degrees\end{tabular}} & \begin{tabular}[c]{@{}c@{}}mm\end{tabular} & \multicolumn{1}{c|}{\begin{tabular}[c]{@{}c@{}}degrees\end{tabular}} & \begin{tabular}[c]{@{}c@{}}mm\end{tabular} & \multicolumn{1}{c|}{\begin{tabular}[c]{@{}c@{}}degrees\end{tabular}} & \begin{tabular}[c]{@{}c@{}}mm\end{tabular} & \multicolumn{1}{c|}{\begin{tabular}[c]{@{}c@{}}degrees\end{tabular}} & \begin{tabular}[c]{@{}c@{}}mm\end{tabular} & \multicolumn{1}{c|}{\begin{tabular}[c]{@{}c@{}}degrees\end{tabular}} & \begin{tabular}[c]{@{}c@{}}mm\end{tabular} & \begin{tabular}[c]{@{}c@{}}degrees\end{tabular} \\ \cline{1-13}

\multicolumn{1}{l|}{\textbf{Hammer} } & \textbf{2.59} $\pm$ \textbf{2.43}  & \multicolumn{1}{|c|}{ \textbf{2.20} $\pm$ \textbf{2.44} }   & \textbf{2.71} $\pm$\textbf{ 2.23}   & \multicolumn{1}{|c|}{ \textbf{2.12} $\pm$ \textbf{2.52} } & \textbf{2.72} $\pm$ \textbf{2.47 }  & \multicolumn{1}{|c|}{ 3.43 $\pm$ 3.21 }   & 5.25 $\pm$ 4.40   & \multicolumn{1}{|c|}{ 2.70 $\pm$ 5.51 }  & 6.37 $\pm$ 5.79   & \multicolumn{1}{|c|}{ 6.66 $\pm$ 5.82 }    &4.14 $\pm$ 3.28 &\multicolumn{1}{|c}{ 3.91 $\pm$ 4.23 } \\

\multicolumn{1}{l|}{\textbf{Screwdriver} }  & \textbf{2.51} $\pm$ \textbf{1.45}  & \multicolumn{1}{|c|}{ \textbf{1.47} $\pm$ \textbf{2.22} }   & 3.07 $\pm$ 2.25   & \multicolumn{1}{|c|}{ 2.84 $\pm$ 3.51 } & \textbf{2.27} $\pm$ \textbf{1.98 }  & \multicolumn{1}{|c|}{ 2.92 $\pm$ 4.13 }   & 4.96 $\pm$ 4.53   & \multicolumn{1}{|c|}{ 3.45 $\pm$ 5.06 }  & 7.19 $\pm$ 5.51   & \multicolumn{1}{|c|}{ 4.22 $\pm$ 3.50 }    &8.74 $\pm$ 4.58 &\multicolumn{1}{|c}{ 9.42 $\pm$ 5.04 } \\

\multicolumn{1}{l|}{\textbf{Bread} }   & \textbf{2.17} $\pm$ \textbf{1.71}  & \multicolumn{1}{|c|}{ \textbf{1.08} $\pm$ \textbf{1.07} }   & 2.98 $\pm$ 2.89   & \multicolumn{1}{|c|}{ \textbf{1.40} $\pm$ \textbf{\textbf{1.10}} } & \textbf{2.32} $\pm$ \textbf{1.74}   & \multicolumn{1}{|c|}{ \textbf{1.54} $\pm$ \textbf{1.72} }   & 3.83 $\pm$ 3.18   & \multicolumn{1}{|c|}{ 2.00 $\pm$ 2.05 }  & 5.91 $\pm$ 5.15   & \multicolumn{1}{|c|}{ 2.14 $\pm$ 2.07 }    &3.70 $\pm$ 3.98 &\multicolumn{1}{|c}{ 2.19 $\pm$ 2.54 } \\

\multicolumn{1}{l|}{\textbf{Spoon} } & {5.05} $\pm${ 4.04 } & \multicolumn{1}{|c|}{ \textbf{1.48 }$\pm$ \textbf{1.70} }   & 7.00 $\pm$ 9.54   & \multicolumn{1}{|c|}{ 4.82 $\pm$ 7.17 } & \textbf{3.13} $\pm$ \textbf{2.90}   & \multicolumn{1}{|c|}{ \textbf{1.45} $\pm$ \textbf{2.14} }   & 46.92 $\pm$ 59.12   & \multicolumn{1}{|c|}{ 15.00 $\pm$ 16.97 }  & 11.11 $\pm$ 12.81   & \multicolumn{1}{|c|}{ 7.01 $\pm$ 9.26 }    &17.37 $\pm$ 17.58 &\multicolumn{1}{|c}{ 11.54 $\pm$ 9.46 } \\

\multicolumn{1}{l|}{\textbf{Wrench} }   & \textbf{1.35} $\pm$ \textbf{1.57}  & \multicolumn{1}{|c|}{ \textbf{1.39} $\pm$ \textbf{2.57} }   & 2.50 $\pm$ 2.09   & \multicolumn{1}{|c|}{ \textbf{1.23} $\pm$ \textbf{1.40} } & \textbf{1.93} $\pm$ \textbf{1.21}   & \multicolumn{1}{|c|}{ \textbf{1.07} $\pm$ \textbf{1.09} }   & 3.82 $\pm$ 2.67   & \multicolumn{1}{|c|}{ \textbf{1.25} $\pm$ \textbf{1.37} }  & 4.47 $\pm$ 3.95   & \multicolumn{1}{|c|}{ 3.88 $\pm$ 4.22 }    &5.43 $\pm$ 3.21 &\multicolumn{1}{|c}{ 2.59 $\pm$ 2.23 } \\

\multicolumn{1}{l|}{\textbf{Glass} }   & \textbf{2.23} $\pm$ \textbf{1.58}  & \multicolumn{1}{|c|}{ \textbf{1.41} $\pm$ \textbf{1.53} }   & 4.79 $\pm$ 6.04   & \multicolumn{1}{|c|}{ 6.48 $\pm$ 7.91 } &\textbf{ 2.96} $\pm$ \textbf{3.03}   & \multicolumn{1}{|c|}{ 3.70 $\pm$ 4.89 }   & 29.46 $\pm$ 30.87   & \multicolumn{1}{|c|}{ 16.27 $\pm$ 16.64 }  & 21.76 $\pm$ 27.29   & \multicolumn{1}{|c|}{ 10.65 $\pm$ 10.51 }    &26.81 $\pm$ 16.01 &\multicolumn{1}{|c}{ 13.87 $\pm$ 10.55 } \\
\cline{1-13} 
\multicolumn{1}{l|}{\textbf{Average} }   & \textbf{2.65} $\pm$ \textbf{1.15}  & \multicolumn{1}{|c|}{ \textbf{1.50} $\pm$ \textbf{0.34} }   & 3.84 $\pm$ 1.60   & \multicolumn{1}{|c|}{ 3.15 $\pm$ 1.91 } &\textbf{ 2.56} $\pm$ \textbf{0.42}   & \multicolumn{1}{|c|}{ 2.35 $\pm$ 1.03 }   & 15.71 $\pm$ 16.69   & \multicolumn{1}{|c|}{ 6.78 $\pm$ 6.31 }  & 9.47 $\pm$ 5.86   & \multicolumn{1}{|c|}{ 5.76 $\pm$ 2.84 }    &11.03 $\pm$ 8.44 &\multicolumn{1}{|c}{ 7.25 $\pm$ 4.57 } \\
\cline{1-13} 
\end{tabular}}

\resizebox{.4\textwidth}{!}{

\color{black}\begin{tabular}{ccccccccccccccc}
& \multicolumn{2}{c}{{\textbf{DINO}}}   & \multicolumn{2}{c}{{\textbf{AspanFormer}}}                                                                                                             
\\ \cline{1-5} 

                                   \multicolumn{1}{c|}{{\textbf{Objects}}} & \begin{tabular}[c]{@{}c@{}}mm\end{tabular} & \multicolumn{1}{c|}{\begin{tabular}[c]{@{}c@{}}degrees\end{tabular}} & \begin{tabular}[c]{@{}c@{}}mm\end{tabular} & \multicolumn{1}{c}{\begin{tabular}[c]{@{}c@{}}degrees\end{tabular}} \\ \cline{1-5}

\multicolumn{1}{l|}{{\textbf{Hammer}} } & 2.61 $\pm$ 3.67  & \multicolumn{1}{|c|}{ 5.16 $\pm$ 3.82 }   & 6.42 $\pm$ 5.59   & \multicolumn{1}{|c}{ 7.13 $\pm$ 7.28 }
 \\

\multicolumn{1}{l|}{{\textbf{Screwdriver}} } & 7.63 $\pm$ 4.64  & \multicolumn{1}{|c|}{ 2.63 $\pm$ 4.18 }   & 6.84 $\pm$ 6.96   & \multicolumn{1}{|c}{ 6.73 $\pm$ 5.39 }  \\

\multicolumn{1}{l|}{{\textbf{Bread}} }& 7.46 $\pm$ 5.81  & \multicolumn{1}{|c|}{ 2.60 $\pm$ 3.01 }   & 15.37 $\pm$ 3.33   & \multicolumn{1}{|c}{ 7.98 $\pm$ 5.22 }  \\

\multicolumn{1}{l|}{{\textbf{Spoon}} }& 16.45 $\pm$ 16.21  & \multicolumn{1}{|c|}{ 9.68 $\pm$ 11.70 }   & 9.64 $\pm$ 15.48   & \multicolumn{1}{|c}{ 8.82 $\pm$ 10.80 }  \\

\multicolumn{1}{l|}{{\textbf{Wrench}} } & 6.85 $\pm$ 3.67  & \multicolumn{1}{|c|}{ 5.25 $\pm$ 5.59 }   & 2.91 $\pm$ 3.13   & \multicolumn{1}{|c}{ 5.36 $\pm$ 4.84 }  \\

\multicolumn{1}{l|}{{\textbf{Glass}} } & 19.13 $\pm$ 18.58  & \multicolumn{1}{|c|}{ 8.08 $\pm$ 8.81 }   & 10.35 $\pm$ 14.80   & \multicolumn{1}{|c}{ 10.52 $\pm$ 11.72 }  \\

\cline{1-5} 
\multicolumn{1}{l|}{{\textbf{Average}} }   & {10.02} $\pm$ {5.80}  & \multicolumn{1}{|c|}{ {5.56} $\pm$ {2.61} }   & 8.59 $\pm$ 3.88   & \multicolumn{1}{|c}{ 7.76 $\pm$ 1.63 }   \\
\cline{1-5}
\end{tabular}}

\end{table*}